\documentclass[11pt]{article}

\usepackage[utf8]{inputenc}
\usepackage[margin=1in]{geometry}
\usepackage{amsmath, amssymb, amsfonts}
\usepackage{graphicx}
\usepackage{booktabs}
\usepackage{xcolor}
\usepackage{microtype}
\usepackage{caption}
\usepackage{subcaption}
\usepackage[hidelinks]{hyperref}
\usepackage[numbers,sort&compress]{natbib}
\usepackage{authblk}
\usepackage{placeins}

\graphicspath{{figs/}{pictures/}{images/}{./}}

\newcommand{\npjFigCaption}[2]{\caption{\textbf{Fig.~\thefigure\ | #1.} #2}}
\newcommand{\npjTableCaption}[2]{\caption{\textbf{Table~\thetable\ | #1.} #2}}

\title{Improving Heart-Focused Medical Question Answering in LLMs via Variance-Aware Rubric Rewards with GRPO}

\author[1,5]{Arash Ahmadi}
\author[2,6]{Parisa Masnadi Khiabani}
\author[1,5]{Sarah Sharif}
\author[3,2]{Charles Nicholson}
\author[4]{David Ebert}
\author[7]{Aswathy K. Vaikom House}
\author[1,5,*]{Mike Banad}

\affil[1]{School of Electrical and Computer Engineering, University of Oklahoma, Norman, OK, USA}
\affil[2]{Data Science and Analytics Institute, University of Oklahoma, Norman, OK, USA}
\affil[3]{School of Industrial and Systems Engineering, University of Oklahoma, Norman, OK, USA}
\affil[4]{Office of Responsible Artificial Intelligence (ORAI), University of Arizona, Tucson, AZ, USA}
\affil[5]{Intelligent Neuromorphic and Quantum Understanding for Innovative Research and Engineering (INQUIRE) Laboratory, University of Oklahoma, Norman, OK, USA}
\affil[6]{Data Institute for Societal Challenges (DISC), University of Oklahoma, Norman, OK, USA}
\affil[7]{University of Oklahoma Health Sciences Center, Oklahoma City, OK, USA}
\affil[*]{Correspondence: \texttt{bana@ou.edu}}
\date{}

\begin{document}
\maketitle

\begin{abstract}
Large language models can support medical question answering, but privacy, cost, and reliability still limit their clinical use in focused domains. Can rubric-based reinforcement learning bring a locally deployable model close to much larger systems? We study Group Relative Policy Optimization (GRPO) for heart-focused medical question answering with rubric supervision from RaR-Medicine. Rather than rely on binary rubric aggregation or a single overall score, our variance-aware reward framework builds continuous rewards from criterion-level outcomes. It preserves partial credit and reduces reward sparsity. We adapt a Qwen3-14B policy with low-rank adapters and quantized weights. On a held-out heart-related HealthBench subset, the best GRPO variant improves accuracy from 0.362 to 0.502 and F1 from 0.532 to 0.668 over the base model, close to GPT-OSS-120B. An independent clinician rated most responses clinically acceptable (61\%) and patient ready (63\%), both above the automated accuracy; fewer than 2\% were unsafe. Taken together, the results support rubric-based reinforcement learning as a practical route for focused medical language models.
\end{abstract}

\noindent
Medical research and clinical care now apply artificial intelligence (AI) to prediction, imaging, and language processing. Predictive AI models support risk stratification and outcome forecasting. In imaging, deep learning systems have demonstrated strong performance on image-based tasks such as skin cancer classification, diabetic retinopathy detection, and breast cancer screening \cite{topol2019high,khalifa2024artificial,esteva2017dermatologist,gulshan2016development,mckinney2020international}. Meanwhile, natural language processing methods are assisting clinicians and researchers in extracting, organizing, and interpreting the growing amount of unstructured text in electronic health records and biomedical literature \cite{huang2019clinicalbert,jerfy2024growing,eguia2024clinical,liu2025registry}. Even so, broader clinical adoption of these techniques relies on reliability, transparency, and the careful handling of sensitive clinical data \cite{topol2019high,thirunavukarasu2023large}.

Large language models (LLMs) continue this trend because they can interpret natural language questions and generate free-form explanations. This capability opens applications in patient triage, in decision-making, and in choosing what kind of explanation a patient should receive. General-purpose LLMs still have trouble with clinical specificity. They can generate plausible sentences that contradict one another, and they can omit any statement of uncertainty. Heart-focused medical question answering makes this challenge concrete. Symptoms such as chest pain, shortness of breath, and palpitations require conservative advice, proper uncertainty management, and risk assessment. The Global Burden of Disease Study 2023~\cite{naghavi2025global} underlines the clinical significance of this domain. Ischemic heart disease and stroke have been the first and second leading causes of age-standardized death globally from 1990 to 2023. Ischemic heart disease held the top spot in every year except 2021, when COVID-19 temporarily took it. This persistent burden makes AI decision support for heart-related clinical reasoning a high priority.

Early clinical natural language processing systems relied on task-specific pipelines. These pipelines extracted entities, retrieved information, applied rule-based heuristics, or used supervised classification. Such methods are controllable and auditable. They require substantial feature engineering, however, and generally do not transfer well to different institutions or styles of notes. Recent open models in medicine, such as Med-Gemma, have demonstrated that domain adaptation can improve factuality and clinical relevance over general baselines \cite{medgemma2025techreport}. The CancerGPT model showed that task-adapted LLMs can provide few-shot biomedical inference in low-data settings \cite{li2024cancergpt}. A lightweight, fine-tuned LLM for symptom-based assessment of depression showed that clinically aligned adaptation can support symptom-level assessment beyond coarse screening \cite{weber2025using}. Work on lightweight disease diagnosis systems likewise points to the need for deployment-aware design in resource-constrained clinical settings \cite{su2025large}. The common lesson is that adapting models, data, and goals to the clinical domain and deployment constraints may be more important than model size.

Recent work on medical LLMs also shows that grounding, retrieval, personalization, explainability, and auditable outputs are all necessary for deployment. The clinical entity augmented retrieval system improves clinical information extraction by retrieving entity-centered note spans rather than relying only on general semantic similarity. This can improve relevance while reducing unnecessary context \cite{lopez2025clinical}. Retrieval augmented generation has also improved local models for radiology contrast media consultation. Coupled with curated knowledge and human oversight, such models can be deployed in a privacy preserving manner \cite{wada2025retrieval}. When patient education agents are built into EHRs, they show the potential of personalized LLM support; they also show the importance of safety controls and clinician oversight in patient-facing applications \cite{hao2025personalizing}. KT-LLM further explores evidence-grounded, policy-aware, and auditable sequence-text modeling \cite{zheng2026kt}. The holistic AI approach in medicine argues that explainability and performance need to be improved together, not treated as independent goals \cite{petridis2026holistic}. These studies present concrete progress toward clinically-grounded and trustworthy LLM systems. None of them, however, directly addresses how to optimize a medical assistant using multi-criterion clinical rubrics post-training.

Domain adaptation often uses supervised fine-tuning (SFT) on curated instruction or dialogue data. Because SFT optimizes imitation of reference answers, it can inherit annotation artifacts and it can encourage memorization of training examples. Neither is desirable when the goal is robust clinical reasoning and safe generalization. SFT also compresses multifactor clinical quality into a single target sequence. Rubric-based evaluation makes this limitation explicit, since clinical answers must satisfy multiple criteria that cover correctness, safety, completeness, and appropriate communication. For high-stakes clinical assistants, a training method that can optimize directly against such criteria is therefore attractive.

Reinforcement learning (RL) provides a complementary paradigm because it optimizes behavior with respect to a reward signal rather than a fixed demonstration target \cite{sutton1998reinforcement}. Early value-based algorithms such as Q-learning established fundamental principles for action-value estimation \cite{watkins1992q}. Subsequent methods such as SARSA and deep Q-networks extended RL to stochastic control and high-dimensional observations \cite{wang2013backward,mnih2013playing}. Double DQN and its subsequent modifications addressing moving-target instability reduced overestimation bias and improved stability \cite{van2016deep,halat2024modified}. Policy-gradient and actor-critic methods also enabled scalable optimization for large action spaces. Advantage actor-critic provided a practical deep RL baseline \cite{babaeizadeh2016reinforcement}. Trust region and proximal policy optimization improved robustness of policy updates \cite{schulman2015trust,schulman2017proximal}. Optimization beyond imitation can yield strategies that are not present in supervised targets; AlphaGo highlighted this capability through self-play and search \cite{silver2017mastering}. AlphaFold extended learning-based optimization to scientific discovery and showed that structured objectives can unlock capabilities that exceed traditional heuristic pipelines in protein structure prediction \cite{jumper2021highly}. More recently, RL post-training for language models has become a practical route to improve reasoning. The DeepSeek-Math and DeepSeek-R1 projects show that RL can raise the performance of relatively small models on difficult reasoning tasks \cite{Shao2024DeepSeekMath,guo2025deepseek}.

Group Relative Policy Optimization (GRPO) is an RL algorithm designed for language model post-training. It avoids an explicit value function and uses group-wise relative advantage estimates \cite{Shao2024DeepSeekMath}. Compared with methods that train a separate critic, this design reduces memory requirements. It also fits reward settings where relative ranking is more stable than absolute calibration. GRPO first showed strong results in mathematical reasoning and has since been extended to a broader set of generative tasks. Recent work shows that GRPO can improve code generation in underrepresented programming languages by integrating reasoning-driven feedback into the optimization loop \cite{pennino2025reasoning}. Related GRPO-based RL has also improved deep reasoning translation, suggesting that reward-driven post-training can transfer beyond domains with exact verifiers into more open-ended generation settings \cite{wang2026deeptrans}. In medical AI, this trend is beginning to appear in multimodal settings. MedVLM-R1 applies reinforcement learning to medical vision-language reasoning, encouraging explicit natural-language reasoning and improving medical reasoning capability in radiological tasks \cite{pan2025medvlm}. RARL similarly combines reinforcement learning, LoRA, and LLM-as-a-judge style evaluation to improve medical vision-language reasoning and generalization under limited data and hardware budgets \cite{pham2025rarl}. These studies strengthen the case for applying GRPO-style post-training to medical dialogue tasks. In such tasks, correctness is multi-dimensional, rewards are sparse, and answer quality depends not only on factual accuracy but also on explanation quality, safety, and completeness.

\begin{figure*}[h] 
  \centering
  \includegraphics[width=\textwidth]{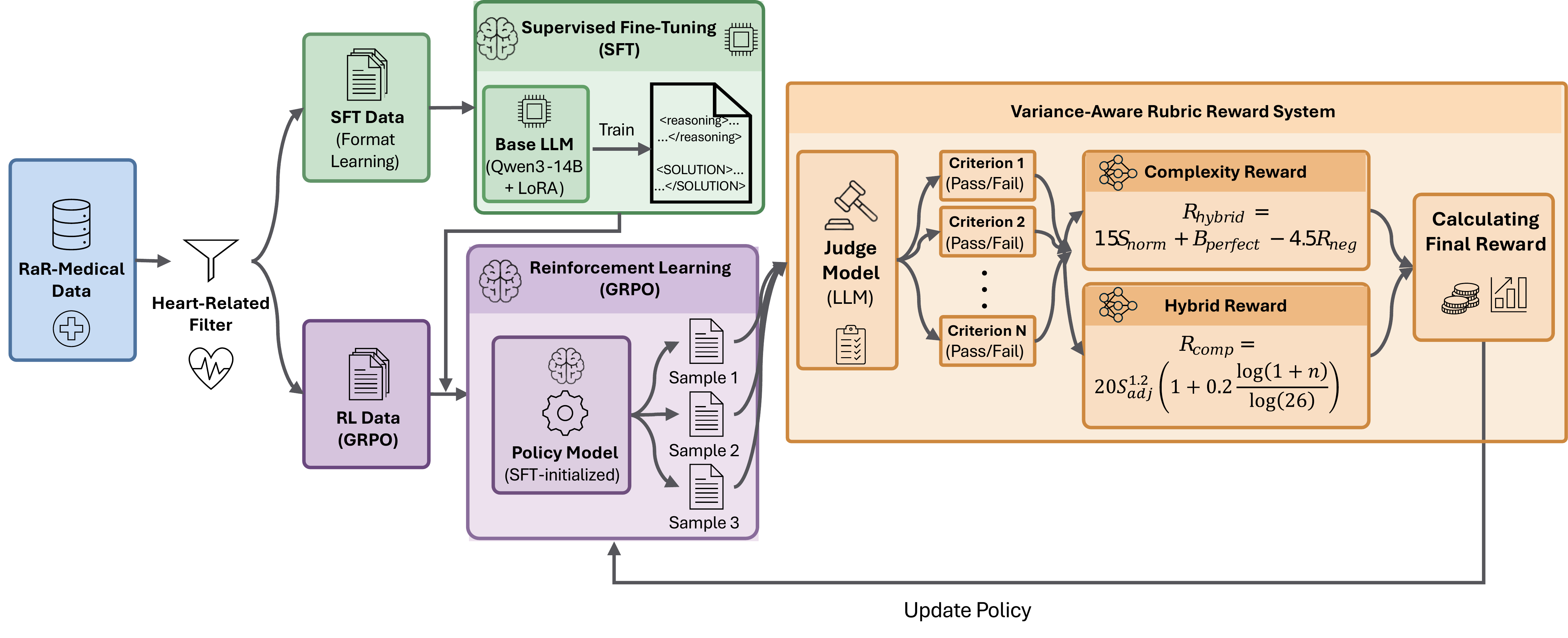}
  \npjFigCaption{Overview of the heart-focused GRPO training and evaluation pipeline}{%
We start from RaR-Medicine data, filter it to heart-related queries, and split it into supervised fine-tuning (SFT) and reinforcement learning (GRPO) subsets. SFT first initializes the base model with structured reasoning outputs. During GRPO, the policy model generates multiple candidate responses per prompt, and an LLM-based judge evaluates them against prompt-specific rubric criteria. Each criterion is scored independently (pass/fail). Variance-aware reward functions, including hybrid and complexity-aware formulations, then aggregate the resulting signals into a scalar reward. Group-relative optimization uses this reward to update the policy, enabling stable learning from multi-criteria clinical feedback.%
}
  \label{fig:framework}
\end{figure*}

In this paper, we develop a GRPO-based framework for a heart-focused medical assistant that answers free-form questions while satisfying rubric-defined clinical criteria. Our training data come from RaR-Medicine. We filter the corpus to queries that are directly related to heart-related problems through a dedicated classifier that uses either an LLM-based decision rule or a high-recall keyword filter. We augment the filtered subset with synthetic reasoning traces generated by MedGemma-27B \cite{medgemma2025techreport} that encourage explicit intermediate explanations. We use a structured output format that separates reasoning from final recommendations through dedicated tags. This choice aligns with prior work on eliciting multi-step reasoning in language models \cite{wei2022chain}. An initial supervised stage teaches the model this format and stabilizes generation. In the main optimization stage, we apply GRPO to a disjoint subset of the same heart-related training pool. Reward computation follows the structure of HealthBench-style rubrics. Each prompt contains positive and negative criteria with associated point values. Figure~\ref{fig:framework} provides an overview of the full pipeline.

A large judge model evaluates each criterion independently and returns a binary decision. This setup follows the broader LLM-as-a-judge direction \cite{zheng2023judging} and the rubrics-as-rewards framework \cite{Gunjal2025Rubrics}. This criterion-wise design is also consistent with recent medical evaluation work showing that expert-grounded automated verification can scale assessment in specialty QA~\cite{giuffre2025expert} and that rubric-like LLM judging can align strongly with human evaluators in clinical summarization~\cite{croxford2025evaluating}. Related medical RL studies such as RARL and broader GRPO-based work such as DeepTrans also support the use of model-based judging and structured reward criteria when exact verifiers are unavailable \cite{pham2025rarl,wang2026deeptrans}. Criterion-level scoring reduces brittleness that arises when a single model assigns an overall score to an entire response. A reward shaping function then transforms the raw rubric scores into a scalar reward. We study multiple shaping families that address sparse rewards, preserve strong incentives for fully correct and safe answers, and account for rubric complexity so that prompts with many criteria still provide meaningful gradients. We focus training on challenging prompts and exclude items with very small rubric sets, because they yield trivial rewards and limited learning signals. In the final system, we adapt a 14B parameter base model with low-rank adapters and quantized weights. This keeps post-training feasible on academic hardware and supports privacy-preserving local deployment.

On a held-out heart-related subset of HealthBench, GRPO post-training improves rubric satisfaction over the same backbone without RL, and the improvements appear across accuracy, F1, recall, and precision. We make four contributions, namely (1)~a rubric-aligned GRPO pipeline for heart-focused medical question answering that supports explicit reasoning traces and structured outputs; (2)~a criterion-wise judging and variance-aware reward shaping strategy that reduces reward sparsity and improves learning under heterogeneous rubrics; (3)~a data curation and filtering pipeline that isolates heart-related questions and produces synthetic reasoning traces for instruction tuning; and (4)~a comparative evaluation against strong baselines that examines the effect of scaling both the judge and the policy models.

\section{Results}

We evaluated rubric based reinforcement learning for clinical reasoning through a series of experiments in cardiac medicine. The analysis covers data curation, model optimization, and final evaluation. We first established a specialized dataset of heart-related inquiries by filtering and restructuring the RaR-Medicine corpus, which ensures high relevance and grading granularity. Following a supervised initialization to stabilize reasoning formats, we applied Group Relative Policy Optimization (GRPO) using three distinct variance-aware reward mechanisms. The sections below describe the characteristics of the curated dataset, the training dynamics of the reward functions, and the post-training performance of the models on the held-out HealthBench evaluation set.

\subsection{Dataset curation and characteristics}
\label{subsec:data}

\subsubsection{Training set: RaR-Medicine with heart-related filtering}
\label{subsubsec:rar}
\begin{figure*}[t]
  \centering
  \includegraphics[width=0.82\textwidth]{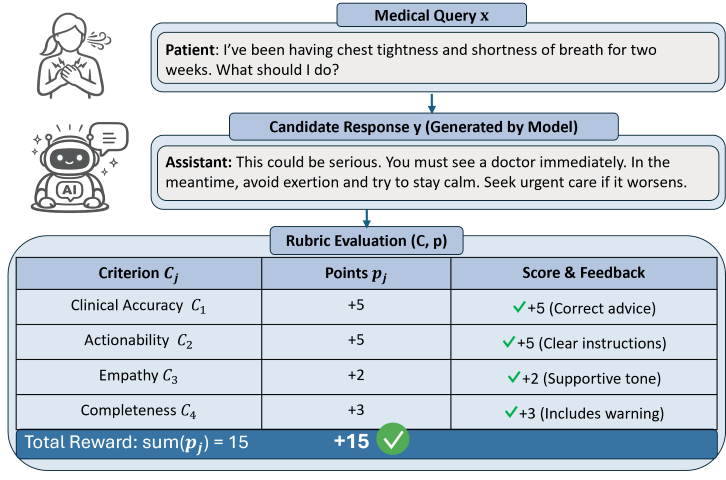}
  \npjFigCaption{Rubric-based supervision format used in our datasets}{%
  A medical query $q$ is paired with a candidate response $y$, which is then evaluated against a prompt-specific rubric $\mathcal{C}(q)=\{(c_j,p_j)\}_{j=1}^m$. Each satisfied criterion contributes its signed point value, and the total example-level score is the sum over satisfied criteria. The example shown is illustrative and is included to clarify how criterion-level rubric annotations are converted into training and evaluation signals.%
  }
  \label{fig:rubric_schema}
\end{figure*}

The RaR-Medicine~\cite{Gunjal2025Rubrics} dataset provides training prompts, reference completions, and rubric annotations. Each example contains a natural-language question, a reference answer, and a rubric set that defines how a completion is graded. The rubric fields are stored as \texttt{criterion} (text), \texttt{points} (signed scalar), and \texttt{title} (optional metadata). The raw dataset comes in Parquet splits. We convert each sample to a JSONL record that contains a chat-style prompt, a reference completion, and a rubric list in which each element contains \texttt{criterion} and \texttt{points}. This conversion produces a schema that is compatible with rubric-based evaluation.
Figure~\ref{fig:rubric_schema} illustrates the rubric-based supervision format used in our data pipeline. For a medical query $q$, a candidate response $y$ is evaluated against a prompt-specific rubric $\mathcal{C}(q)=\{(c_j, p_j)\}_{j=1}^m$, where each criterion $c_j$ describes a clinically meaningful desired or undesired behavior and $p_j$ denotes its signed point value. Summing the points of all satisfied criteria gives the total example-level score. In RaR-Medicine, the reference completion provides the supervised target during fine-tuning. Later, we use the same rubric structure to evaluate generated responses and derive reward signals for reinforcement learning.
We restrict training to heart-related samples. A dedicated classifier assigns a binary label \texttt{heart\_related} and auxiliary metadata that includes a theme category and keyword evidence. The classifier queries an instruction-following medical model and parses a structured decision from the model output. We use MedGemma~\cite{sellergren2025medgemma} in this role because it is specialized for medical instruction following and its compact footprint supports large-scale preprocessing under limited computational resources. As a result, we included a \texttt{synthetic\_reasoning} field that contains a model-generated reasoning trace aligned with the reference answer. 

Table~\ref{tab:dataset_example} shows an illustrative training instance after preprocessing. It follows the format of the JSONL files and shows how rubric criteria describe both required and prohibited behaviors.

\begin{table}[t]
\centering
\small
\begin{tabular}{p{0.22\linewidth} p{0.72\linewidth}}
\toprule
Field & Example value \\
\midrule
Prompt $q$ & \texttt{What should I do if someone is experiencing chest pain?} \\
Reference completion $o^{\star}$ & \texttt{Call emergency services immediately. Seek urgent evaluation if chest pain is severe, new, or associated with shortness of breath, profuse perspiration, or pain that radiates to the arm or jaw.} \\
Rubric $\mathcal{C}(q)$ & \texttt{Advises immediate emergency evaluation for possible acute coronary syndrome} ($w=2$); \texttt{Mentions calling emergency services} ($w=2$); \texttt{Avoids providing unsafe medication instructions} ($w=-2$); \texttt{Mentions red-flag symptoms that increase urgency} ($w=1$). \\
\bottomrule
\end{tabular}
\npjTableCaption{Illustrative example of a rubric-annotated instance in our training format}{The criteria and weights shown here are representative and are included to clarify the data schema.}
\label{tab:dataset_example}
\end{table}

\subsubsection{Train split usage}
\label{subsubsec:split}

After filtering, we shuffle the heart-related training subset with a fixed random seed and split it into two disjoint halves. One half is used for supervised fine-tuning; the other half is reserved for reinforcement learning with Group Relative Policy Optimization. This split eliminates overlap between stages, and it isolates the effect of reward optimization.

\begin{figure*}[p]
  \centering
  \includegraphics[width=\textwidth]{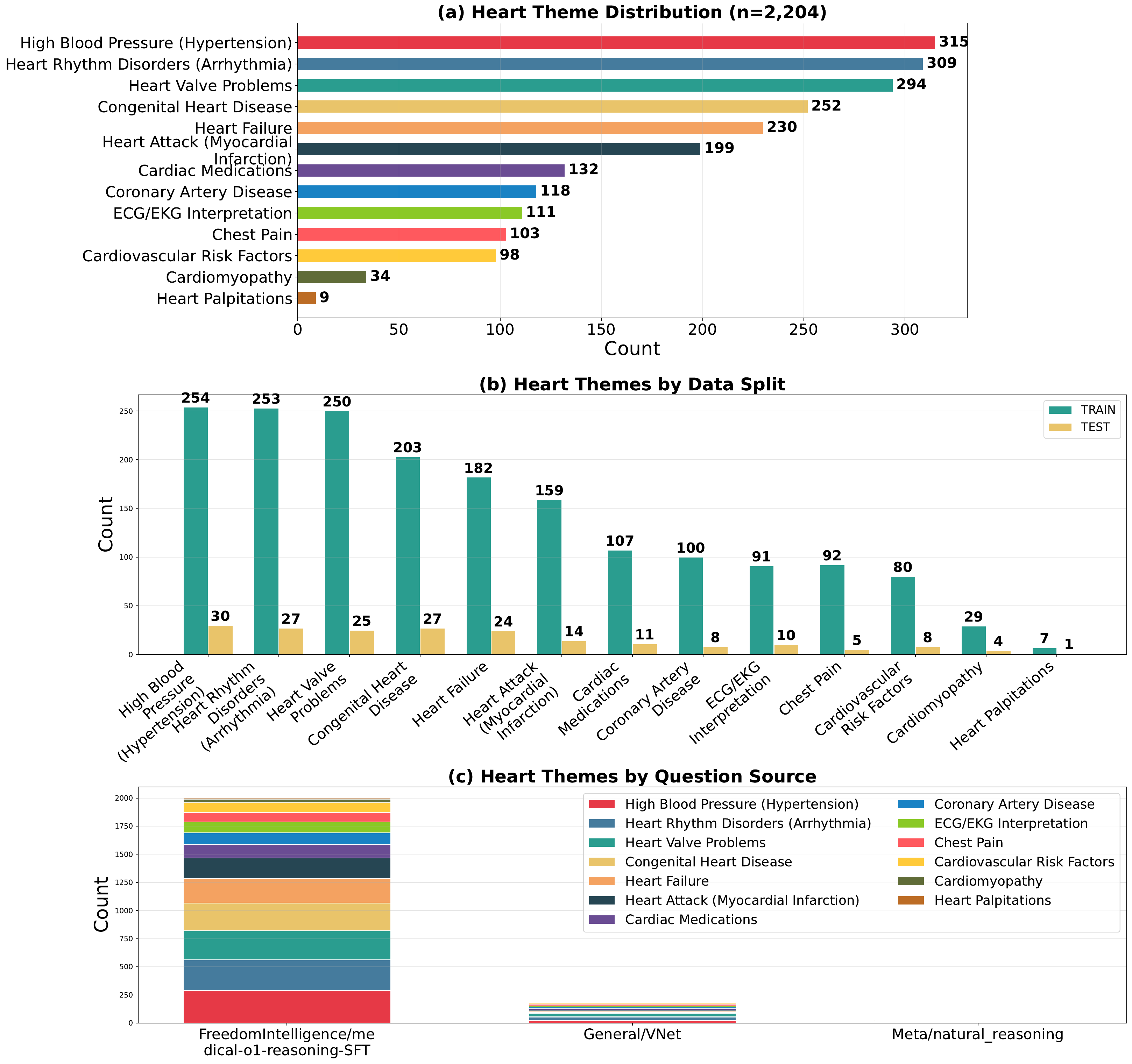}
  \npjFigCaption{Dataset composition of the filtered heart-related RaR-Medicine subset}{%
    (a)~Distribution of heart-related themes (excluding ``Other'' and themes with fewer than five records).
    (b)~Theme counts by dataset split.
    (c)~Theme distribution stratified by question source.%
  }
  \label{fig:dataset_overview}
\end{figure*}

\paragraph{Dataset summary statistics.}
Figure~\ref{fig:dataset_overview} shows the breakdown of the filtered heart-related subset of the RaR-Medicine dataset, including overall theme distribution and split and source stratification. Supplementary Fig.~4 shows the heart-related vs.\ non-heart-related sample balance, and Supplementary Fig.~5 presents the question source distribution. Additional descriptive statistics on rubric counts (Supplementary Figs.~6--7), question and answer lengths (Supplementary Figs.~8--9, 11--12), and rubric weight distributions (Supplementary Fig.~10) are provided in the Supplementary Information.

\subsubsection{Evaluation set: HealthBench}
\label{subsubsec:healthbench}

We evaluate on HealthBench~\cite{arora2025healthbench} and treat it as held out. HealthBench contains 5,000 multi-turn health conversations with physician-written rubric criteria created by 262 physicians across 26 medical specialties. We evaluate on a held-out, non-synthetic subset of HealthBench. Why HealthBench? It focuses on medical question answering and has standard rubrics, which makes it well-suited for held-out evaluation. Its scope is also not too large for us to repeatedly compute metrics during development. Our evaluation pipeline filters to \texttt{heart\_related = YES} and computes Accuracy, Precision, Recall, and F1 against available physician-derived binary labels. The reported results use $n=500$ heart-related evaluation examples sampled with seed~42.

\subsection{Model performance on the held-out HealthBench heart subset}
\label{subsec:main_performance}

Table~\ref{tab:model_comparison} reports the performance of the primary benchmark models on the held-out heart-related subset of HealthBench. Extended multi-metric comparisons, radar charts, performance heatmaps, and category comparisons are provided in Supplementary Figs.~15--18. Supplementary Video~1 shows the cumulative accuracy of the primary benchmark models as evaluation progresses sample by sample across the 500 held-out prompts. Among all systems, Kimi-K2, which features approximately 1 trillion parameters, achieves the highest overall performance with an accuracy of 0.570 and F1 score of 0.726. GPT-OSS-120B follows with an accuracy of 0.508 and F1 score of 0.674. 

Notably, our locally trained GRPO-optimized Qwen3 variants achieve performance on par with the much larger GPT-OSS-120B. The GRPO (COMPLEXITY) reward reaches an accuracy of 0.502 and F1 score of 0.668, while the GRPO (HYBRID) reward yields an accuracy of 0.498 and F1 score of 0.665. For comparison, the Qwen3-14B Base model achieves an accuracy of 0.362 and F1 score of 0.532. These results demonstrate that variance-aware reward shaping effectively improves model performance. Kimi-K2 does achieve the highest overall performance. Even so, it is an open-source model whose approximately 1 trillion total parameters far exceed the memory capacity of academic-grade hardware such as the NVIDIA RTX 6000 PRO, which makes local training or serving infeasible. By contrast, we train and serve our GRPO variants entirely on a single workstation GPU. This demonstrates that rubric-based RL can close a substantial fraction of the gap to frontier-scale models under strict hardware constraints.

\begin{table*}[!t]
\centering
\small
\begin{tabular}{l c c c}
\toprule
Model & Accuracy & 95\% CI & F1 \\
\midrule
Kimi-K2 & \textbf{0.570} & [0.527, 0.613] & \textbf{0.726} \\
GPT-OSS-120B & 0.508 & [0.464, 0.552] & 0.674 \\
Qwen3-14B GRPO (COMPLEXITY) & 0.502 & [0.458, 0.546] & 0.668 \\
Qwen3-14B GRPO (HYBRID) & 0.498 & [0.454, 0.542] & 0.665 \\
MedGemma-27B & 0.448 & [0.404, 0.492] & 0.619 \\
Gemma3-12B & 0.442 & [0.398, 0.486] & 0.613 \\
Phi4-14B & 0.442 & [0.398, 0.486] & 0.613 \\
Llama-4-Scout-17B & 0.432 & [0.389, 0.475] & 0.603 \\
Qwen3-14B GRPO (RaR-IMPLICIT) & 0.412 & [0.369, 0.455] & 0.584 \\
Llama-3.3-70B & 0.398 & [0.355, 0.441] & 0.569 \\
Qwen3-14B GRPO (RaR-EXPLICIT) & 0.396 & [0.353, 0.439] & 0.567 \\
MedGemma-4B & 0.396 & [0.353, 0.439] & 0.567 \\
Llama-4-Maverick-17B & 0.390 & [0.347, 0.433] & 0.561 \\
Qwen3-14B Base & 0.362 & [0.320, 0.404] & 0.532 \\
MedGemma-1.5-4B & 0.322 & [0.281, 0.363] & 0.487 \\

\bottomrule
\end{tabular}
\npjTableCaption{Model performance comparison on the held-out heart-related HealthBench subset ($n=500$, seed~42)}{Models are sorted by Accuracy. Variance-aware GRPO reward functions substantially improve the Qwen3-14B base model, while external frontier models such as Kimi-K2 and GPT-OSS-120B achieve the highest overall performance.}
\label{tab:model_comparison}
\end{table*}

\begin{figure*}[tp]
  \centering
  \vspace{-10pt} 
  \includegraphics[width=\textwidth, keepaspectratio]{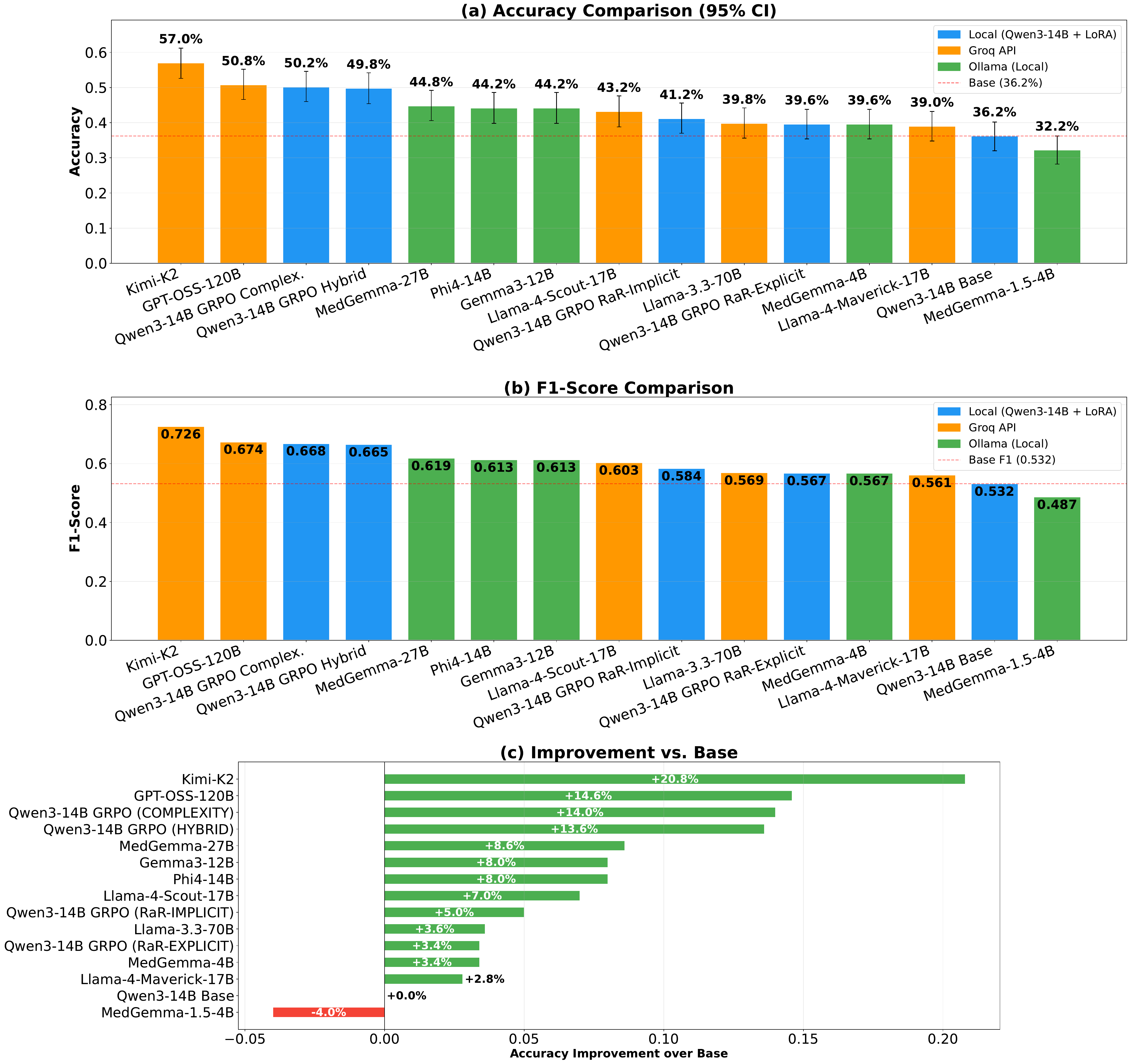}
  \npjFigCaption{Main performance on the held-out heart-related HealthBench subset ($n=500$, seed~42)}{%
    (a)~Accuracy comparison across the primary benchmark models with 95\% confidence intervals. Kimi-K2 achieves the highest accuracy (0.570), followed by GPT-OSS-120B and the GRPO-trained Qwen3 variants.
    (b)~F1-score comparison showing similar ranking trends to accuracy.
    (c)~Accuracy improvement ($\Delta$Accuracy) relative to the Qwen3-14B Base model. Variance-aware GRPO reward functions (COMPLEXITY and HYBRID) achieve the largest gains among local deployments.%
  }
  \label{fig:main_performance}
\end{figure*}

Table~\ref{tab:improvement} reports improvements relative to the Qwen3-14B Base model. Figure \ref{fig:main_performance} illustrates the benchmark performance, including accuracy with confidence intervals, F1-score comparison, and relative accuracy improvement. The GRPO (COMPLEXITY) and GRPO (HYBRID) local models show the highest improvements.

\begin{table}[!t]
\centering
\small
\begin{tabular}{l c c c c}
\toprule
Model & Acc $\Delta$ & Acc $\Delta$\% & F1 $\Delta$ & F1 $\Delta$\% \\
\midrule
GRPO (COMPLEXITY) & +0.140 & +38.7\% & +0.137 & +25.7\% \\
GRPO (HYBRID) & +0.136 & +37.6\% & +0.133 & +25.0\% \\
GRPO (RaR-IMPLICIT) & +0.050 & +13.8\% & +0.052 & +9.8\% \\
GRPO (RaR-EXPLICIT) & +0.034 & +9.4\% & +0.036 & +6.7\% \\
\bottomrule
\end{tabular}
\npjTableCaption{Performance improvements relative to the Qwen3-14B Base model ($n=500$)}{Variance-aware reward functions (COMPLEXITY and HYBRID) produce substantially larger gains than the RaR-based reward strategies.}
\label{tab:improvement}
\end{table}

\FloatBarrier

\subsection{Ablations over reward shaping, judge scale, and training stability}
\label{subsec:ablations}

\begin{figure*}[p]
  \centering
  \includegraphics[width=0.9\textwidth]{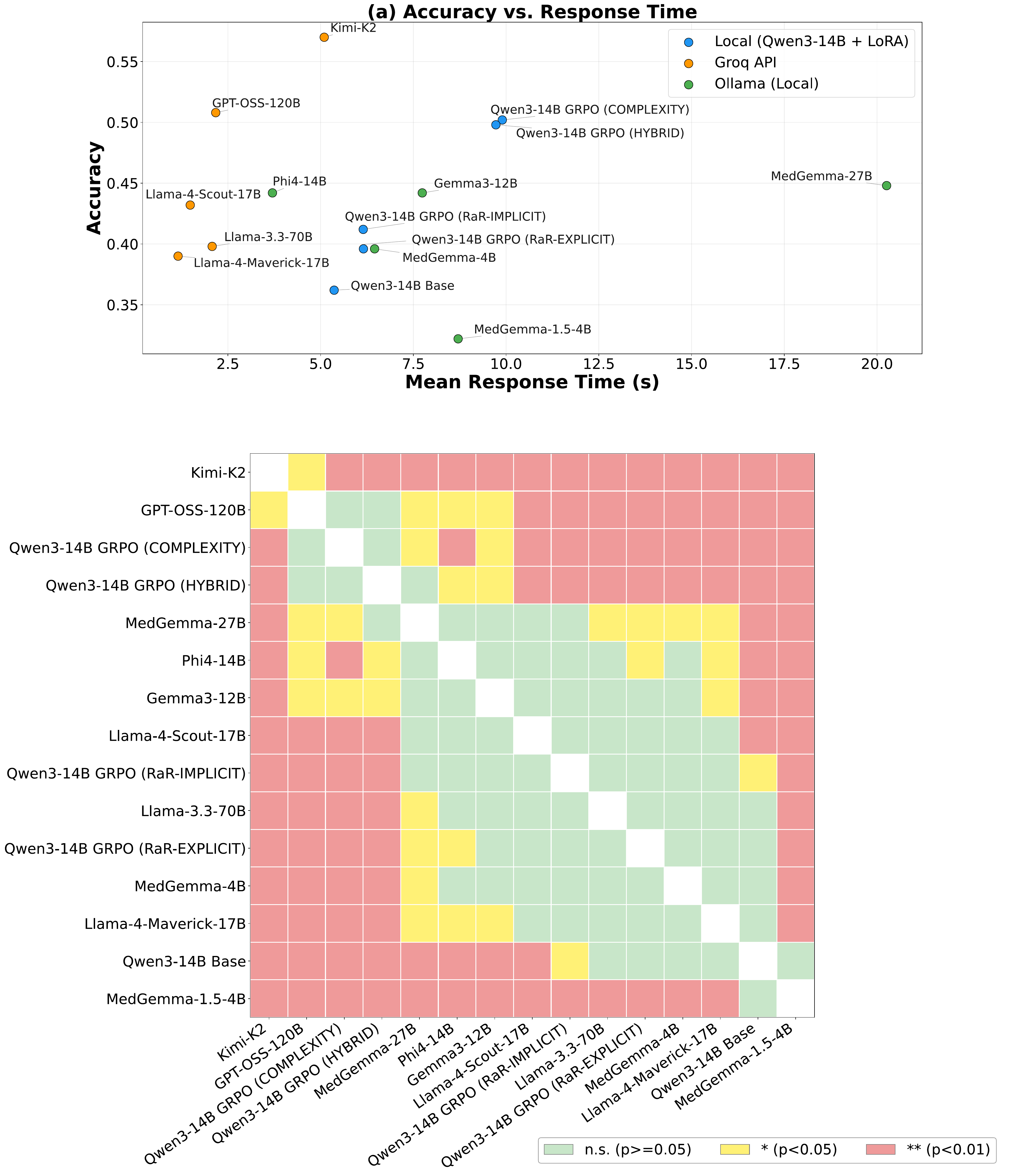}
  \npjFigCaption{Pairwise McNemar significance and average response time}{%
    Left: pairwise McNemar tests for statistical significance of prediction differences; cells report not significant (n.s.), significant ({*}, $p<0.05$), or highly significant ({**}, $p<0.01$).
    Right: mean response time (in seconds) for each primary benchmark model, characterizing the latency dimension of the performance-deployment tradeoff. Additional per-model response time details are provided in Supplementary Fig.~19.%
  }
  \label{fig:ablation_deployment}
\end{figure*}

Figure~\ref{fig:ablation_deployment} presents pairwise McNemar significance tests alongside mean response times across the primary benchmark models. Supplementary Video 2 presents a per-sample, per-criterion comparison of the Base, GRPO (Complexity), GPT-OSS-120B, and MedGemma-27B models, illustrating which criteria are satisfied and which are missed for each evaluation prompt.

We additionally trained two GRPO variants using the reward aggregation strategies proposed in the original Rubrics as Rewards (RaR) framework~\cite{Gunjal2025Rubrics}. The first, RaR-Explicit, independently evaluates each rubric criterion via an LLM judge and aggregates binary satisfaction signals through a normalized weighted sum with fixed categorical weights (\textit{Essential}: 1.0, \textit{Important}: 0.7, \textit{Optional}: 0.3, \textit{Pitfall}: 0.9). The second, RaR-Implicit, passes all criteria holistically to the judge and elicits a single Likert score normalized to $[0,1]$. Both variants produced only modest improvements over the Qwen3-14B base model, with +9.4\% and +13.8\% relative accuracy, respectively. Those gains fell far short of the +38.7\% and +37.6\% gains achieved by the Complexity and Hybrid rewards (Table~\ref{tab:improvement}). The difference is statistically significant. Pairwise McNemar tests confirm that both our variance-aware rewards outperform RaR-Explicit ($p < 10^{-5}$) and RaR-Implicit ($p < 10^{-3}$).

Why did the RaR aggregation strategies help so little in our setting? We attribute their limited effectiveness to three factors. First, the explicit aggregation relies on rigid, hand-tuned categorical weights that impose a fixed importance hierarchy across all prompts; this one-size-fits-all mapping cannot adapt to the heterogeneous complexity of cardiac medicine queries, where the relative salience of individual criteria varies substantially by clinical context. Second, the implicit aggregation delegates the entire scoring decision to a single holistic LLM judgment. It collapses the multi-dimensional rubric information into a coarse Likert score that discards granular criterion-level signal and introduces judge-level variance. Third, neither RaR strategy accounts for rubric complexity; each treats a prompt with five criteria identically to one with eighteen. A base model can often satisfy all criteria on simple rubrics approximately out of the box, so these easy prompts contribute little discriminative training signal. The harder and more informative case arises when rubric sets are large. Satisfying seventeen out of eighteen criteria on a complex prompt represents a substantially greater achievement than a perfect score on a five-criterion prompt. Yet both RaR aggregation strategies assign comparable normalized rewards to both outcomes. Our variance-aware reward functions address this asymmetry explicitly. The Complexity-aware variant applies a logarithmic bonus that scales with rubric size, so that high satisfaction on demanding prompts produces a stronger reward signal and therefore a larger policy gradient. This design amplifies learning from precisely the prompts where the base model struggles most. It converts partial-credit differences on complex rubrics into high-value training signal that neither the fixed-weight explicit aggregation nor the holistic implicit scoring can provide.

\begin{figure*}[tp]
  \centering
  \includegraphics[width=\textwidth]{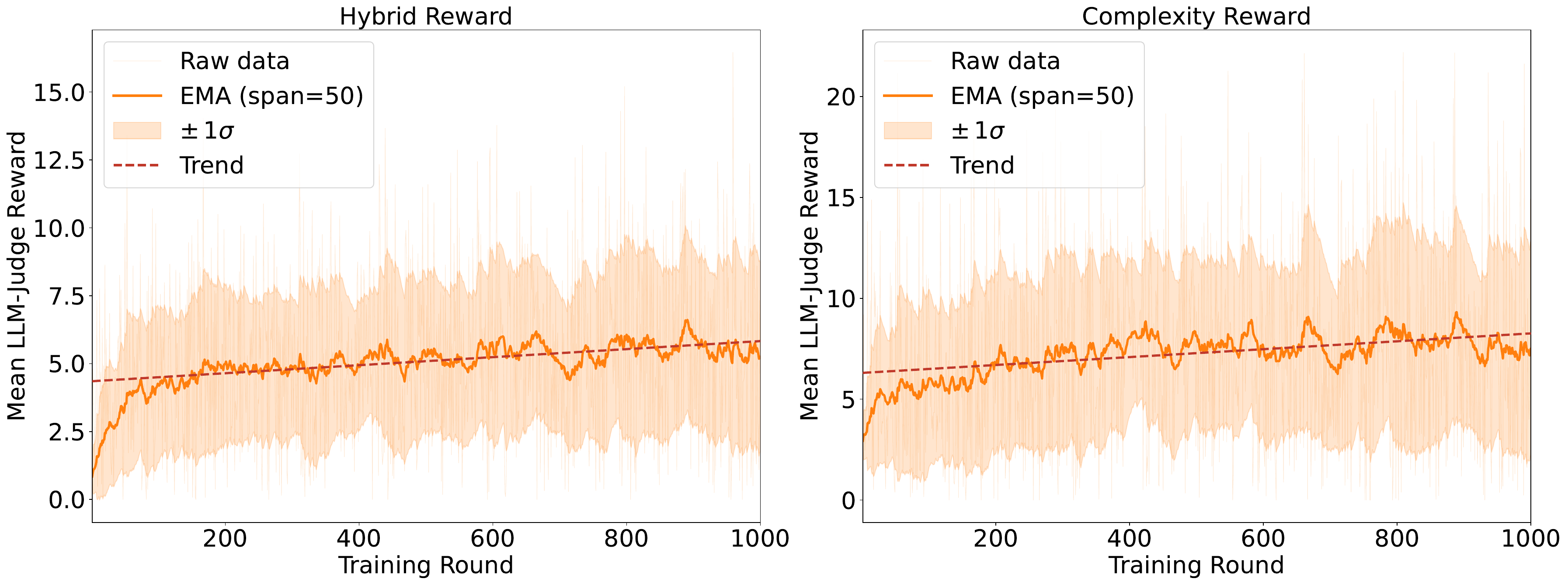}
  \npjFigCaption{Training reward dynamics for the Hybrid and Complexity reward functions}{%
    Mean LLM-judge reward over 1000 GRPO training rounds. Light lines show the raw per-step reward; the bold curve shows an exponential moving average (EMA, span\,=\,50); the shaded band indicates $\pm 1\sigma$ around the EMA; the dashed line shows the linear trend. Both reward functions show steady improvement over the course of training.%
  }
  \label{fig:reward_trends}
\end{figure*}

Figure~\ref{fig:reward_trends} shows the mean LLM-judge reward over 1000 GRPO training rounds for the Hybrid and Complexity reward functions. Both curves display substantial per-step variance. In GRPO, this variance is expected and structurally beneficial. Group-relative advantage estimation normalizes rewards within each sampled group by subtracting the group mean and dividing by the group standard deviation~\cite{Shao2024DeepSeekMath}, so that reward spread across completions is converted into differential advantage signals that drive policy improvement. The exponential moving average (EMA) and the linear trend show a steady improvement in rewards throughout training. The $\pm 1\sigma$ band illustrates the per-step reward variance that arises from stochastic prompt sampling and group-level generation. Notably, the Complexity reward operates in a higher absolute range, consistent with its rubric-size scaling described above, while the Hybrid reward shows a tighter variance envelope.

\FloatBarrier

\subsection{Clinical validation by expert review}
\label{subsec:clinical_validation}

Automated rubric scoring can identify the correctness of responses at a criterion level. It cannot alone be used to verify whether responses are safe, appropriately hedged, and useful at the bedside. To validate the automated metric against expert judgment, we asked a practicing clinician to independently review the deployment model, namely the GRPO (Complexity) variant that is our strongest local configuration. The reviewer assessed the model responses to 56 held-out heart-related prompts with a structured instrument. This instrument scored seven dimensions on a 1 to 5 Likert scale (clinical correctness, completeness, safety, emergency referral appropriateness, context awareness, patient communication, and actionability). It also recorded binary judgments of patient readiness and usability without editing, and categorical ratings of acceptability and harm severity. The reviewed prompts spanned all nine clinical strata of the evaluation set, namely acute coronary syndrome, arrhythmia and syncope, hypertension, heart failure, valvular disease and cardiomyopathy, prevention of coronary artery disease, cardiac medications, and ECG interpretation. This review is a preliminary validation by a single reviewer. An expanded assessment with multiple reviewers, led by the clinical coauthors, is ongoing.

Among the seven graded dimensions, the model scored highest on patient communication ($4.05 \pm 1.03$) and clinical correctness ($3.84 \pm 1.16$), followed by actionability ($3.75 \pm 1.24$), completeness ($3.59 \pm 1.30$), emergency referral appropriateness ($3.53 \pm 1.41$, scored on the 32 cases where referral was applicable), and safety ($3.52 \pm 1.35$). Context awareness was the weakest dimension ($2.64 \pm 1.65$). This score captures the documented tendency of language models to answer directly without expressing uncertainty or asking for missing clinical context~\cite{yin2023know,zhou2024relying}. These scores are summarized in Table~\ref{tab:clinical_validation}.

At the response level, the clinician rated 60.7\% of responses clinically acceptable (34 of 56, Wilson 95\% CI 47.6 to 72.4\%), and 76.8\% acceptable or acceptable with only minor issues (43 of 56, 64.2 to 85.9\%). A majority were judged patient ready (62.5\%, 35 of 56, 49.4 to 74.0\%) and usable without editing (58.9\%, 33 of 56, 45.9 to 70.8\%). Only a single response was rated unsafe (1.8\%, 1 of 56, 0.3 to 9.4\%), and 23.2\% were flagged as unsafe or otherwise problematic (13 of 56, 14.1 to 35.8\%). On the harm severity scale, 62.5\% of responses carried no or only minor risk of harm, while 14.3\% (8 of 56, 7.4 to 25.7\%) were rated as serious or life threatening if acted upon without further clinical input. The most frequent failure modes noted by the reviewer were failure to solicit clinical context (9 cases), overly broad disclaimers (5 cases), missed emergency referral (5 cases), and false reassurance (4 cases). These failures concentrate in context gathering rather than factual error. That pattern is consistent with the low context awareness score and with only one incorrect fact flagged in all 56 cases.

The acceptability rate reported by the clinician (60.7\%) and the patient readiness rate (62.5\%) both exceed the model's automated HealthBench accuracy for the same configuration (0.502, Table~\ref{tab:model_comparison}).

\begin{table}[!t]
\centering
\small
\begin{tabular}{l c}
\toprule
Evaluation dimension & Mean $\pm$ SD \\
\midrule
Patient communication & $4.05 \pm 1.03$ \\
Clinical correctness & $3.84 \pm 1.16$ \\
Actionability & $3.75 \pm 1.24$ \\
Completeness & $3.59 \pm 1.30$ \\
Emergency referral appropriateness$^{\dagger}$ & $3.53 \pm 1.41$ \\
Safety & $3.52 \pm 1.35$ \\
Context awareness & $2.64 \pm 1.65$ \\
\bottomrule
\end{tabular}
\npjTableCaption{Clinician ratings of the GRPO (Complexity) model on 56 held-out heart-related prompts}{Each dimension was scored on a 1 to 5 Likert scale by an independent practicing clinician (higher is better). $^{\dagger}$Emergency referral appropriateness was scored only on the 32 cases where an emergency referral decision was applicable. Response level acceptability, patient readiness, safety, and harm severity summaries are reported in the text.}
\label{tab:clinical_validation}
\end{table}

\FloatBarrier

\section{Discussion}
What counts as a `correct' answer in medicine? The difficulty of defining it often hinders the switch from Supervised Fine-Tuning to Reinforcement Learning in healthcare. Unlike similar reinforcement learning tasks in games like AlphaGo \cite{granter2017alphago} or protein folding algorithms like AlphaFold \cite{jumper2021highly}, clinical diagnosis lacks a simulator. This work highlights a subtle but critical failure mode when adapting techniques like Rubrics as Rewards (RaR) \cite{Gunjal2025Rubrics} to algorithms like GRPO. Optimization algorithms relying on batch normalization are intolerant of sparse, binary rewards.

These results echo challenges reported in recent digital medicine literature \cite{thirunavukarasu2023large}, where generalist models fail to capture nuance. A ``soft'' reward signal that acknowledges partial correctness, such as fulfilling 17 out of 18 criteria, helps stabilize training and aligns the model's incentives with the iterative process of clinical reasoning.

The supervised tuning step prepares the output format for GRPO. It trains the model to output the required reasoning and answer tags reliably, which is necessary for the downstream reward computation. Without this initial stage, the base Qwen3-14B model does not consistently produce the structured output format necessary for rubric evaluation. Once the outputs are well formed, GRPO introduces a rubric aligned optimization phase that directly maximizes criterion satisfaction. The performance improvements in Table \ref{tab:model_comparison} therefore indicate the effectiveness of the full pipeline, from the base model through the two training stages.

The LLM judge in our pipeline operates at the individual criterion level. For each rubric criterion, it receives the prompt context and the model completion and returns a binary ``present'' decision with a short justification. This task is narrower than holistic scoring approaches, in which a single model assigns an overall score for the quality of the entire response. The judge uses low temperature, JSON-constrained output, and a retry policy, all of which reduce noise and improve reproducibility. The original Rubrics as Rewards framework~\cite{Gunjal2025Rubrics} demonstrated that GPT-4o-mini can perform this criterion-level matching task effectively. Our pipeline uses GPT-OSS-120B as the judge model instead. It ranks higher on the Chatbot Arena leaderboard~\cite{chiang2024chatbot} (Elo 1354 vs.\ 1317 for GPT-4o-mini) and provides stronger overall capability for nuanced criteria.

We designed the final adapted policy model for local deployment. With 4-bit quantization and LoRA adapters, the 14B parameter model can be run on a single workstation GPU such as the NVIDIA RTX 6000 PRO. This supports privacy preserving inference without transmitting patient data to external services. For training and evaluation, we serve the judge model (GPT-OSS-120B) through a Groq-hosted inference backend, which keeps judging latency per criterion feasible in an academic setting. GPT-OSS-120B is an open source model, and it should be possible to run it locally on sufficiently large hardware. We use Groq because of inference speed and academic hardware limitations, not because of access to the model.

This work targets heart-focused medical question answering as a clinically motivated testbed for rubric-aligned reinforcement learning. The scope is intentional. Cardiovascular disease represents the leading cause of death globally~\cite{naghavi2025global}, and the associated clinical reasoning demands conservative guidance, appropriate uncertainty handling, and careful risk assessment. Recent work in \textit{npj Digital Medicine} has shown that narrow clinical scope is productive for evaluating AI systems, with published studies focusing on radiology contrast-media consultation~\cite{wada2025retrieval}, symptom-based depression scoring~\cite{weber2025using}, and prostate-cancer patient education~\cite{hao2025personalizing}. Our contribution is methodological. We show that variance-aware reward shaping improves rubric satisfaction on a clinically relevant task, and that reward design principles may generalize to other rubric-based medical QA domains.

Several directions remain for future investigation. First, the automated judging is corroborated by a preliminary clinical validation with a single reviewer (Section~\ref{subsec:clinical_validation}); a larger prospective study with multiple reviewers would allow estimates of agreement between reviewers and strengthen clinical validation further. Second, extending our pipeline to other clinical domains beyond medical QA for the heart would put the generality of variance-aware reward shaping to the test. Finally, although our policy model can be deployed locally, lowering the computational cost of criterion-level judgment during training would make our method more accessible to resource-limited research groups.

\section{Methods}
\label{sec:methods}

Here we describe the datasets, preprocessing steps, model configuration, training process, and reward design that we used to train a medical assistant on heart related questions using rubric based supervision. We begin by converting medical prompts annotated with rubrics into a standard chat format. A short supervised phase stabilizes the response structure. Reinforcement learning then optimizes a continuous variance aware reward based on rubric judgments for each criterion.

\subsection{Terminology and notation}
\label{subsec:terminology}

A \emph{prompt} is a user query and is denoted by $q \in \mathcal{Q}$, where $\mathcal{Q}$ is the set of prompts.
A \emph{completion} is the sequence of tokens generated by a language model in response to $q$ and is denoted by $o = (o_1,\dots,o_{|o|})$, where each $o_t$ is a token from a tokenizer vocabulary $\mathcal{V}$ and $|o|$ is the completion length.

A \emph{policy model} is the language model interpreted as a stochastic decision rule over tokens.
The policy model is parameterized by $\theta$ and defines a conditional distribution over completions through an autoregressive factorization,
\begin{equation}
\pi_{\theta}(o \mid q) = \prod_{t=1}^{|o|}\pi_{\theta}(o_t \mid q, o_{<t}),
\end{equation}
where $o_{<t}=(o_1,\dots,o_{t-1})$.

Each prompt $q$ is paired with a rubric set $\mathcal{C}(q)=\{(c_k,w_k)\}_{k=1}^{C}$, where $c_k$ is a natural-language criterion, $w_k \in \mathbb{Z}$ is its point value, and $C$ is the number of criteria.
Positive weights encode desirable properties that the completion should satisfy.
Negative weights encode undesirable properties that the completion should avoid.
A \emph{judge model} evaluates a completion against each criterion and returns a binary decision.
A \emph{reward function} maps the criterion-level decisions into a scalar reward $r \in \mathbb{R}$.

\subsection{Model and output format with parameter-efficient adaptation}
\label{subsec:model}

\subsubsection{Base model and parameter-efficient adaptation}
\label{subsubsec:base_model}
We initialize the policy from Qwen3-14B-Base. To reduce memory usage, we train with 4-bit quantization and apply Low-Rank Adaptation to a subset of attention and feed-forward projection matrices. For a weight matrix $W_0 \in \mathbb{R}^{d_{\mathrm{out}} \times d_{\mathrm{in}}}$, Low-Rank Adaptation parameterizes the adapted weight as
\begin{equation}
W = W_0 + \Delta W, \qquad \Delta W = BA,
\end{equation}
where $A \in \mathbb{R}^{r \times d_{\mathrm{in}}}$, $B \in \mathbb{R}^{d_{\mathrm{out}} \times r}$, and $r$ is the Low-Rank Adaptation rank. We set $r=16$ and use a Low-Rank Adaptation scaling factor $\alpha_{\mathrm{LoRA}} = 2r$. We apply adaptation to the attention projections $\{q_{\mathrm{proj}}, k_{\mathrm{proj}}, v_{\mathrm{proj}}, o_{\mathrm{proj}}\}$ and the feed-forward projections $\{\mathrm{gate}_{\mathrm{proj}}, \mathrm{up}_{\mathrm{proj}}, \mathrm{down}_{\mathrm{proj}}\}$.

\subsubsection{Response structure and chat template}
\label{subsubsec:format}
The assistant response follows a two-part format that separates a reasoning trace from the final answer. The reasoning segment is delimited by \texttt{<start\_working\_out>} and \texttt{<end\_working\_out>}. The solution segment is delimited by \texttt{<SOLUTION>} and \texttt{</SOLUTION>}. A fixed system message instructs the model to generate outputs in this format. The chat template of the tokenizer starts generating from the reasoning start token. We introduce this formatting during supervised fine-tuning and preserve it during reinforcement learning.

\subsection{Training procedure}
\label{subsec:training}

\subsubsection{Supervised Fine-Tuning warm start}
\label{subsubsec:sft}
Supervised fine-tuning provides a format warm start for reinforcement learning. Before training, the base model does not reliably emit the required tags, which makes downstream parsing and reward computation ambiguous. This step teaches the model to place its reasoning trace and its final answer in the correct places and it creates a stable response structure that is maintained throughout reinforcement learning.

We convert each supervised example into a chat transcript with a system message, a user message, and an assistant message. The assistant message concatenates a reasoning trace and the reference completion inside the required tags. The learning objective maximizes the likelihood of the reference completion under the policy model, conditioned on the prompt,
\begin{equation}
\mathcal{L}_{\mathrm{SFT}}(\theta)
=
-\mathbb{E}_{(q,o^{\star})}\left[\sum_{t=1}^{|o^{\star}|}\log \pi_{\theta}(o^{\star}_t \mid q, o^{\star}_{<t})\right].
\end{equation}

We restrict loss computation to the assistant response tokens. Tokens that belong to the system and user messages receive a mask value of $-100$ so that they do not contribute to the gradient. We also filter the supervised dataset by transcript length and drop examples whose tokenized transcripts exceed 90\% of the maximum sequence length.

\subsubsection{Group Relative Policy Optimization post-training}
\label{subsubsec:grpo}
For reinforcement learning post-training, we use Group Relative Policy Optimization~\cite{Shao2024DeepSeekMath}. GRPO is especially well suited for rubric-based medical question answering, since it does not require training a separate value network, and instead computes advantages using group-wise relative rewards. A key implication of this design is that learning depends on within-group reward variation. If all sampled completions for a prompt receive the same reward, the normalized advantages collapse toward zero and the policy receives little or no learning signal. A good reward function for GRPO should therefore retain partial-credit information and lead to non-trivial dispersion among completions of varying quality.

For each prompt $q$, the old policy $\pi_{\theta_{\mathrm{old}}}$ samples a group of $G$ outputs $\{o_i\}_{i=1}^{G}$. The objective maximized by Group Relative Policy Optimization is
\begin{equation}
\label{eq:group_relative_policy_optimization_objective}
\begin{aligned}
\mathcal{J}_{\mathrm{GRPO}}(\theta)
&=
\mathbb{E}_{q \sim P(Q), \{o_i\}_{i=1}^{G} \sim \pi_{\theta_{\mathrm{old}}}(O \mid q)}
\Bigg[
\frac{1}{G}\sum_{i=1}^{G}\frac{1}{|o_i|}\sum_{t=1}^{|o_i|}
\Bigg(
\min\Big[
\rho_{i,t}(\theta)\,\widehat{A}_{i,t},
\operatorname{clip}\!\left(\rho_{i,t}(\theta), 1-\epsilon, 1+\epsilon\right)\widehat{A}_{i,t}
\Big]
\\
&\hspace{70pt}
-\beta_{\mathrm{KL}}\,\mathbb{D}_{\mathrm{KL}}\!\left[\pi_{\theta}\,\|\,\pi_{\mathrm{ref}}\right]
\Bigg)
\Bigg],
\end{aligned}
\end{equation}
where $\epsilon>0$ is the clipping parameter and $\beta_{\mathrm{KL}}\ge 0$ is the coefficient of the Kullback-Leibler regularization term~\cite{kullback1951information}. The likelihood ratio is
\begin{equation}
\rho_{i,t}(\theta) = \frac{\pi_{\theta}(o_{i,t}\mid q,o_{i,<t})}{\pi_{\theta_{\mathrm{old}}}(o_{i,t}\mid q,o_{i,<t})}.
\end{equation}

We use outcome-level supervision, so the advantage is constant across tokens within a completion:
\begin{equation}
\widehat{A}_{i,t} = \widetilde{r}_i \quad \text{for all } t \in \{1,\dots,|o_i|\}.
\end{equation}
The normalized group reward $\widetilde{r}_i$ is computed from the raw rewards $\{r_i\}_{i=1}^{G}$ as
\begin{equation}
\widetilde{r}_i = \frac{r_i - \operatorname{mean}(\{r_j\}_{j=1}^{G})}{\operatorname{std}(\{r_j\}_{j=1}^{G})}.
\end{equation}

We filter the reinforcement learning dataset by rubric count and prompt length. We remove prompts whose rubric sets have fewer than four criteria, because prompts with short rubrics tend to generate trivial reward patterns and weaker group-wise discrimination. We also compute the 90th percentile of prompt token lengths and retain prompts at or below this threshold. Our experiments use $G=6$ sampled completions per prompt and a maximum completion length of 1024 tokens.

\subsection{Rubric-based reward computation}
\label{subsec:reward}

\subsubsection{Reward design principles}
\label{subsubsec:reward_principles}
The optimization requirements of GRPO guide our reward design. Because GRPO normalizes rewards within a sampled group, a useful reward function should satisfy four properties. First, it should produce non-zero variance whenever completions differ in quality. Second, it should be monotonic, so that better rubric performance receives higher reward. Third, it should maintain information from the rubric assessment instead of collapsing partial credit into a binary result. Fourth, it should be complexity-aware, since prompts with more criteria typically represent harder evaluation problems than prompts with only a few rubric items.

\subsubsection{Criterion-level judging}
\label{subsubsec:criterion_judging}
We evaluate each rubric criterion independently with a separate judge model. The judge model takes the prompt context and model completion as inputs, and outputs a decision in the form of a structured JSON object with a binary \texttt{present} field and a short justification. We query the judge once per criterion.

Let $m_k \in \{0,1\}$ indicate whether the judge marks criterion $c_k$ as present in the completion. We aggregate positive and negative contributions separately. The achieved positive score and the maximum possible positive score are
\begin{equation}
s^{+} = \sum_{k: w_k>0} w_k m_k,
\qquad
s^{+}_{\max} = \sum_{k: w_k>0} w_k.
\end{equation}
The achieved negative magnitude and the maximum possible negative magnitude are
\begin{equation}
s^{-} = \sum_{k: w_k<0} |w_k| m_k,
\qquad
s^{-}_{\max} = \sum_{k: w_k<0} |w_k|.
\end{equation}

We define the normalized positive score and normalized negative ratio as
\begin{equation}
s_{\mathrm{norm}} = \frac{s^{+}}{\max(s^{+}_{\max},1)},
\qquad
\rho =
\begin{cases}
\frac{s^{-}}{\max(s^{-}_{\max},1)} & \text{if } s^{-}_{\max}>0,\\
0 & \text{if } s^{-}_{\max}=0.
\end{cases}
\end{equation}
We also define two indicators for exact positive satisfaction and zero negative violations:
\begin{equation}
\mathbb{I}_{\mathrm{all\ pos}} = \mathbb{I}[s^{+} \ge s^{+}_{\max}],
\qquad
\mathbb{I}_{\mathrm{no\ neg}} = \mathbb{I}[s^{-} = 0].
\end{equation}

\subsubsection{General reward formulation}
\label{subsubsec:general_reward_formulation}
Let $n_c = |\mathcal{C}(q)|$ denote the number of rubric criteria for prompt $q$, and let $n_{\max}$ be the maximum rubric count in the training set. In our data, $n_{\max}=25$. We define a general parametric reward family
\begin{equation}
\label{eq:general_reward}
r = r_{\mathrm{base}} \cdot \hat{s}^{\,\alpha}
\cdot
\left(
1 + \beta \cdot \frac{\log(1+n_c)}{\log(1+n_{\max})}
\right),
\end{equation}
where $r_{\mathrm{base}}$ is a base reward scale, $\alpha>0$ controls the curvature of the reward function, and $\beta\ge 0$ controls the strength of the complexity bonus.

The effective score $\hat{s}$ incorporates both positive credit and negative-criteria penalties:
\begin{equation}
\label{eq:effective_score}
\hat{s} = \max\!\left(0,\; s_{\mathrm{norm}} - \lambda \rho \right),
\end{equation}
where $\lambda \ge 0$ is the penalty coefficient for negative criteria. This formulation preserves partial credit, penalizes unsafe or undesirable content, and ensures that low-quality completions do not receive inflated rewards.

\subsubsection{Hyperparameter selection and justification}
\label{subsubsec:reward_hyperparameter_selection}
We selected the reward hyperparameters as theory-informed design constants rather than learned parameters. They aim to satisfy four constraints imposed by GRPO training, namely non-trivial reward variance within each sampled group, monotonicity with respect to rubric quality, preservation of partial credit information, and mild awareness of rubric complexity.

Both reward variants use a base reward scale of 20. This choice keeps rewards in a numerically stable and interpretable range for GRPO while matching the scale used throughout the training implementation. Within that shared range, the Hybrid reward allocates 15 points to a continuous base term and 5 points to a perfection bonus. The 15-point base ensures that partially correct responses still receive informative gradients, whereas the 5-point bonus creates a clear but not overwhelming incentive for completions that satisfy all positive criteria and avoid all negative criteria.

For the Hybrid penalty term, the code applies a maximum subtraction equal to 30\% of the 15-point base, which yields the coefficient 4.5. This penalty is strong enough to discourage unsafe or incomplete answers, but not so aggressive that most partially correct completions collapse to zero reward. In the Complexity-aware reward, we chose the exponent $\alpha=1.2$ to mildly sharpen preference for high-quality completions relative to a linear mapping without making the reward too sparse for intermediate outputs. This exponent therefore encourages movement from good to excellent responses while preserving useful gradients for partial progress.

The complexity coefficient $\beta=0.2$ provides only a modest logarithmic bonus as the rubric size increases. This acknowledges that prompts with more criteria are typically harder, but it keeps answer quality as the dominant signal. The negative penalty strength $\lambda=0.5$ reduces the effective score before exponentiation, so harmful outputs are penalized both directly and through the nonlinear transform, increasing separation among low- and mid-quality completions. Finally, the criterion-count normalization uses $n_{\max}=25$, matching the rubric scale assumed by the implementation, so that complexity adjustments remain bounded and comparable across prompts.

\subsubsection{Reward variants}
\label{subsubsec:reward_variants}
We implement and compare two reward variants derived from the general formulation.

\paragraph{Complexity-aware reward.}
The first variant directly instantiates Eq.~\ref{eq:general_reward} with $r_{\mathrm{base}}=20$, $\alpha=1.2$, $\beta=0.2$, $\lambda=0.5$, and $n_{\max}=25$:
\begin{equation}
\label{eq:complexity_reward}
r_{\mathrm{complexity}}
=
20\,\hat{s}^{\,1.2}
\left(
1 + 0.2\frac{\log(1+n_c)}{\log(26)}
\right),
\end{equation}
where
\begin{equation}
\hat{s} = \max\!\left(0,\; s_{\mathrm{norm}} - 0.5\rho\right).
\end{equation}
This reward emphasizes high scores through the power transform, incorporates an explicit bonus for prompts with larger rubric sets, and applies the negative-criteria penalty before the nonlinear transformation so that harmful outputs are penalized more strongly. In practice, the range is approximately $[0,25]$, with slight overshoot possible for highly complex prompts that achieve very strong rubric performance.

\paragraph{Hybrid reward.}
The second variant separates reward into a continuous base component and a discrete perfection bonus:
\begin{equation}
\label{eq:hybrid_reward}
r_{\mathrm{hybrid}}
=
\max\!\left(0,\; B\,s_{\mathrm{norm}} - 0.3B\,\rho\right)
+
P\,\mathbb{I}\!\left[\mathbb{I}_{\mathrm{all\ pos}}=1 \ \wedge\ \mathbb{I}_{\mathrm{no\ neg}}=1\right],
\end{equation}
where $B=15$ is the base component and $P=5$ is the perfection bonus. This construction yields a linear, interpretable reward for partial success while creating a clear jump at complete positive satisfaction with no negative violations. The proportional negative penalty reduces reward for harmful or incomplete responses without automatically forcing the reward to zero. The output range is $[0,20]$.

The two variants emphasize different aspects of learning. The Complexity-aware reward is better aligned with prompts whose rubric sets are large and heterogeneous, while the Hybrid reward provides a simpler continuous signal together with a strong discrete incentive for flawless responses. Both preserve partial credit, and both avoid the information loss that would result from using a binary pass-fail reward.

\subsubsection{GRPO compatibility of the reward functions}
\label{subsubsec:grpo_reward_compatibility}
We designed both proposed reward variants to remain compatible with GRPO. First, they produce continuous outputs, which makes non-zero reward variance much more likely within a sampled group. Second, they are monotonically non-decreasing in rubric quality, so higher positive scores and fewer negative violations lead to larger rewards. Third, they retain information from criterion-level judgments instead of reducing rubric outcomes to a single binary label. Fourth, the Complexity-aware variant explicitly incorporates rubric size through the logarithmic complexity bonus, while the Hybrid variant retains difficulty information indirectly through the normalized criterion-level scores. These properties make the reward signal more informative for policy optimization than sparse binary aggregation.

\subsection{Implementation details}
\label{subsec:implementation_details}
All experiments use a maximum sequence length of 4096 tokens. We load the base model with 4-bit quantization, and training updates only Low-Rank Adaptation parameters with rank $r=16$ on the attention projections and the feed-forward projections. We use gradient checkpointing to reduce activation memory and fix the random seed to 3407 for data shuffling and sampling.

\begin{figure*}[tp]
  \centering
  \includegraphics[width=\textwidth]{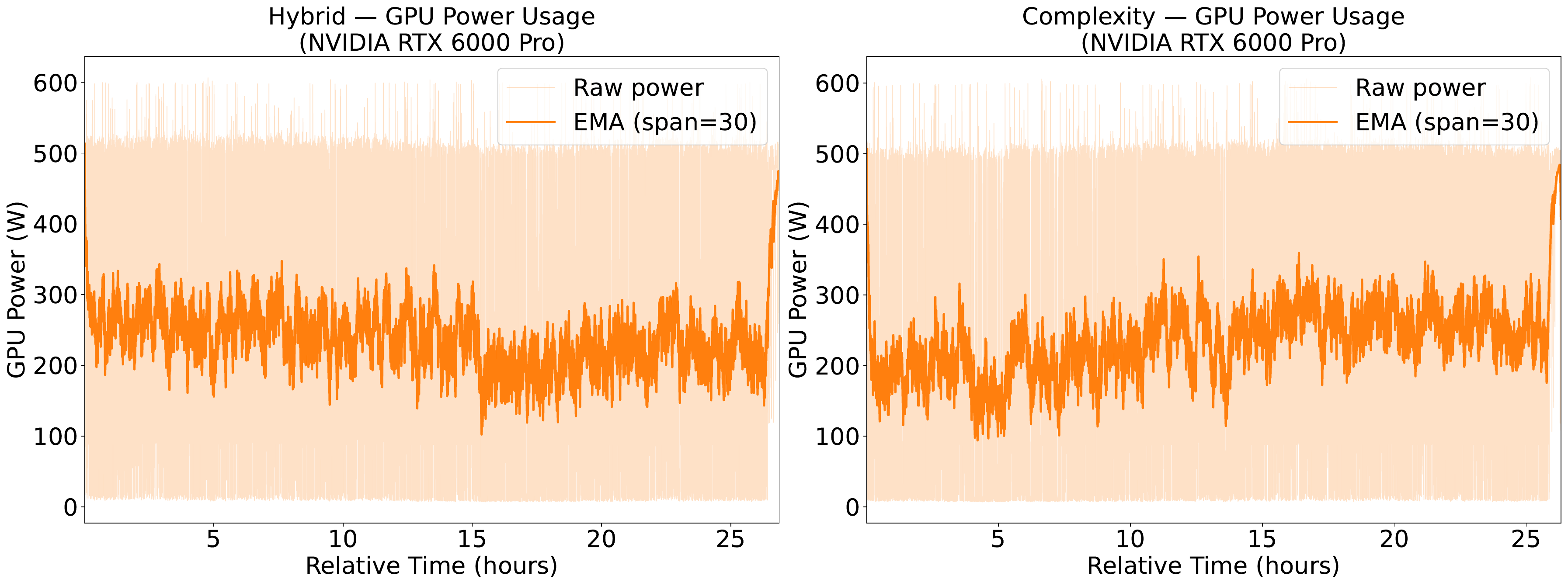}
  \npjFigCaption{GPU power consumption during GRPO training on the NVIDIA RTX 6000 PRO}{%
    GPU power draw (in watts) over the full training duration for the Hybrid (left) and Complexity (right) reward functions. Light lines show raw 15-second power samples; the bold curve shows an exponential moving average (EMA, span\,=\,30). The NVIDIA RTX 6000 PRO (Blackwell Workstation Edition) has a maximum rated power of 600\,W; both runs sustain an average draw of 200--300\,W, with periodic spikes during compute-intensive backward passes.%
  }
  \label{fig:gpu_power}
\end{figure*}

\paragraph{Hardware and GPU utilization.}
We execute all training runs on a single NVIDIA RTX 6000 PRO (Blackwell Workstation Edition) GPU, which has a maximum power consumption of 600\,W. Figure~\ref{fig:gpu_power} shows the GPU power draw over the duration of each GRPO training run. Both the Hybrid and Complexity reward runs sustain approximately 200--300\,W on average during active training, with brief spikes reaching the full 600\,W envelope during peak backward-pass computation. Each reward-function training run takes approximately 26 hours. Where does that time go? Criterion-level judging dominates the wall-clock time, because each rubric contains multiple criteria and each criterion requires an individual call to the LLM judge. The actual model update itself is a lightweight operation.

\paragraph{Inference backend for the judge model.}
To mitigate the latency bottleneck of criterion-level judging, we use the Groq inference platform~\cite{moon2024latency} to serve the judge model. Groq provides substantially lower per-call latency than local GPU inference, which is critical because each training step requires multiple independent judge evaluations per completion per criterion. Without a fast inference backend, the cumulative judge latency would extend each 26-hour training run by an impractical margin. Groq also frees the local GPU completely for policy model training and generation using vLLM. We chose Groq purely for inference speed for the judge. Models with very large parameter counts, such as Kimi-K2 with approximately 1 trillion total parameters, cannot be served locally on the RTX 6000 PRO, and we evaluate them through their respective API endpoints.

Supervised fine-tuning uses a per-device batch size of 4 and gradient accumulation of 4, which yields an effective batch size of 16. Optimization uses AdamW in 8-bit mode with learning rate $2\times 10^{-4}$, weight decay $10^{-3}$, and a linear learning rate schedule with 5 warmup steps. Training runs for two epochs with a cap of 500 update steps, and it logs every step.

Reinforcement learning post-training uses one prompt per update step and samples $G=6$ completions per prompt. Optimization uses AdamW in 8-bit mode with learning rate $5\times 10^{-6}$, weight decay $10^{-2}$, and a linear learning rate schedule with warmup ratio 0.1. Sampling uses temperature 1.0 and a decoding configuration that sets $\texttt{min\_p}=0.1$, $\texttt{top\_p}=1.0$, and $\texttt{top\_k}=-1$. The maximum completion length is 1024 tokens. To compare reward variants under the same conditions, we save Low-Rank Adaptation checkpoints at fixed step intervals in the training loop.

\subsection{Evaluation methodology and metrics}
\label{subsec:evaluation_methods}
We evaluate on HealthBench and treat it as held out. HealthBench contains medical prompts and clinician-derived quality signals and supports rubric-based evaluation by associating prompts with explicit grading criteria. In our pipeline, we merge the HealthBench prompt file with a separate rubric file using \texttt{prompt\_id}. We filter the evaluation set to heart-related prompts and compute Accuracy, Precision, Recall, and F1 with respect to available physician-provided binary labels. All reported results are computed on $n=500$ heart-related evaluation examples sampled with a fixed random seed of 42.

\subsection{Statistical analysis}
\label{subsec:stats}
Accuracy confidence intervals use the normal approximation for a binomial proportion, $\hat{p} \pm 1.96\sqrt{\hat{p}(1-\hat{p})/n}$, with $n=500$. This approximation is appropriate for all reported models because $n\hat{p}$ and $n(1-\hat{p})$ both exceed 5. We assess pairwise differences in model correctness with two sided McNemar tests on paired predictions from the same 500 evaluation examples. The test statistic is $\chi^2=(n_{10}-n_{01})^2/(n_{10}+n_{01})$, where $n_{10}$ and $n_{01}$ denote discordant pairs. We use $p<0.05$ as the nominal threshold for statistical significance. For the 17 model configurations considered in the broader evaluation, there are $\binom{17}{2}=136$ possible pairwise tests; the main comparisons of GRPO variants against the base model remain significant under Bonferroni correction. Exact test statistics, $p$ values, and significance labels are reported in Supplementary Note~7. In the clinician review (Section~\ref{subsec:clinical_validation}), we report the response level proportions with Wilson score 95\% confidence intervals. For the small sample and the proportions near zero or one in that analysis, these intervals are more appropriate than the normal approximation.

\subsection{Ethical considerations}
\label{subsec:ethics}
This study uses publicly available benchmark datasets (RaR-Medicine and HealthBench) that do not contain identifiable patient information. RaR-Medicine provides synthetic and curated medical question-answer pairs with rubric annotations. HealthBench provides physician-authored rubrics grounded in realistic but synthetically generated health conversations~\cite{arora2025healthbench}. No human subjects were recruited, no identifiable patient data were collected or processed, and no clinical interventions were performed as part of this research. The study therefore did not require Institutional Review Board (IRB) approval or informed consent. All model outputs are intended for research evaluation and are not designed for direct clinical use without further validation and clinician oversight.

\subsection{AI use declaration}
\label{subsec:ai_use}
During manuscript preparation, we used generative AI for language editing and to improve the flow of selected passages. The authors developed the research ideas, algorithms, experimental design, and scientific interpretations. The figures and reported numerical results were not generated or changed by these language editing tools. The authors retain responsibility for the accuracy and final wording of the manuscript.

Large language models were used as part of the training and evaluation workflow. MedGemma-27B generated synthetic reasoning traces for supervised fine-tuning examples, as described in the training procedure and Supplementary Note~3. GPT-OSS-120B served as the criterion-level judge during reward computation and evaluation, as described in the rubric-based reward computation section and Supplementary Note~1. The authors designed the study, wrote and reviewed the manuscript, verified the analyses, and take full responsibility for the final content.

\section*{Data availability}
The source training data are derived from the publicly available RaR-Medicine dataset~\cite{Gunjal2025Rubrics}. The evaluation data are derived from the publicly available HealthBench benchmark~\cite{arora2025healthbench}. The processed heart-related dataset used for this study (Supplementary Data~1, \texttt{Supplementary\_Data\_1\_heart\_related\_training\_data.jsonl}) and the anonymized clinician evaluation ratings that support the expert review validation in Section~\ref{subsec:clinical_validation} (Supplementary Data~2, \texttt{Supplementary\_Data\_2\_clinician\_evaluation.json}) are available at \url{https://github.com/INQUIRELAB/variance-aware-rubric-rewards-grpo}.

\section*{Code availability}
The training, evaluation, and reward computation code (Supplementary Code~1, \texttt{Supplementary\_Code\_1\_Hybrid\_Complexity\_Qwen3\_14B.py}) and two supplementary videos are available at \url{https://github.com/INQUIRELAB/variance-aware-rubric-rewards-grpo}. Supplementary Video~1 animates the cumulative accuracy of the primary benchmark models across the 500 held-out heart-related HealthBench samples, and Supplementary Video~2 visualizes per-criterion satisfaction for the Base, GRPO (Complexity), GPT-OSS-120B, and MedGemma-27B models on each evaluation prompt.

\section*{Acknowledgements}
The authors thank Dr.\ Aswathy K.\ Vaikom House for conducting the independent clinical evaluation of model outputs that underpins the expert review validation reported in Section~\ref{subsec:clinical_validation}. The authors acknowledge the use of the Groq inference platform for serving the judge model during training and evaluation, and the open-source Unsloth framework for parameter-efficient fine-tuning. Computational resources were provided by the Inquire Lab at the University of Oklahoma. No specific external funding was received for this work.

\section*{Author contributions}
A.A.\ and P.M.K.\ conceived the study, designed the experiments, implemented the training pipeline, analyzed the results, and drafted the manuscript. S.S.\ contributed to data curation, clinical framing, and manuscript revision. C.N.\ contributed to methodology, statistical interpretation, and manuscript revision. D.E.\ contributed to responsible AI framing and manuscript revision. A.K.V.H.\ performed the independent clinical evaluation of model outputs and contributed to the clinical validation analysis. M.B.\ supervised the project, guided experimental design, and revised the manuscript. All authors approved the final manuscript. A.A.\ and P.M.K.\ are co-first authors and contributed equally.

\section*{Competing interests}
The authors declare no competing interests.

\section*{Additional information}
The Supplementary Information (Supplementary Notes, Figures, and Tables) is appended after the references of this preprint. Code, data, and videos are available at \url{https://github.com/INQUIRELAB/variance-aware-rubric-rewards-grpo}.
Correspondence and requests for materials should be addressed to M.B.\ (\texttt{bana@ou.edu}).

\bibliographystyle{naturemag}
\bibliography{references}

\end{document}


\maketitle

\tableofcontents
\newpage

\section{Supplementary Note 1: Exact Prompts and Templates}
\label{supp:sec:prompts}

This section provides the verbatim prompt texts and template structures used in training and evaluation. For design rationale, see the Methods section of the main paper.

\subsection{System prompt}

\begin{lstlisting}[caption={System prompt prepended to every conversation during SFT and GRPO.}]
You are a helpful medical AI assistant specializing in
heart-related health questions.
When responding to medical queries, think through the problem
carefully and provide your reasoning.
Place your reasoning and analysis between <start_working_out>
and <end_working_out>.
Then, provide your final response and recommendations between
<SOLUTION></SOLUTION>
\end{lstlisting}

\subsection{Structured output format}

\begin{lstlisting}[caption={Two-part output structure enforced through SFT and preserved during GRPO.}]
<start_working_out>
[Reasoning trace: clinical analysis, differential
 considerations, evidence synthesis]
<end_working_out>
<SOLUTION>
[Final clinical recommendation or answer]
</SOLUTION>
\end{lstlisting}

\subsection{Chat template structure}

The tokenizer concatenates a system message, user message, and assistant prefix. When the generation prompt flag is enabled, the template appends the \texttt{<start\_working\_out>} tag so that every generated response begins in the reasoning section.

\begin{lstlisting}[caption={Simplified chat template.}]
{system_message}<|endoftext|>
{user_message}<|assistant|>
<start_working_out>
\end{lstlisting}

\subsection{LLM-as-judge prompt template}

The following prompt is sent to the judge model once per rubric criterion. The variables \texttt{\{question\}}, \texttt{\{response\_text\}}, and \texttt{\{criterion\}} are substituted at run time.

\begin{lstlisting}[caption={LLM-as-judge prompt for criterion-level evaluation.}, label=lst:judge_prompt]
Judge whether the following [response] to [question] is
correct or not based on the precise and unambiguous
[correct_answer] below.

[question]: {question}

[response]: {response_text}

Your judgement must be in the format and criteria
specified below:

extracted_final_answer: The final exact answer extracted
from the [response]. Put the extracted answer as 'None'
if there is no exact, final answer to extract from the
response.

[correct_answer]: {criterion}

reasoning: Explain BRIEFLY (1-2 sentences) why the
extracted_final_answer is correct or incorrect based on
[correct_answer]. Focus ONLY on whether the answers match.
Be concise.

correct: Answer 'yes' if extracted_final_answer matches
the [correct_answer] given above, or is within a small
margin of error for numerical problems. Answer 'no'
otherwise, i.e. if there is any inconsistency, ambiguity,
non-equivalency, or if the extracted answer is incorrect.

confidence: The extracted confidence score between 0% and
100% from [response]. Put 100 if there is no confidence
score available.

Respond with ONLY valid JSON in this exact format:
{
  "present": "yes" or "no",
  "justification": "brief explanation"
}
\end{lstlisting}

The judge response is validated against a Pydantic schema that requires a binary \texttt{present} field and a free-text \texttt{justification} field. Up to 3 retries are attempted per criterion; if all fail, the criterion is treated as not met (\texttt{present = "no"}).

\section{Supplementary Note 2: Heart-Related Classification Details}
\label{supp:sec:classification}

The main paper (Results~\S1.1) describes the LLM-based classifier (MedGemma) and mentions a keyword-based fallback. This section provides the complete keyword dictionary used by the fallback classifier.

\begin{table}[h]
\centering
\small
\begin{tabular}{l p{10cm}}
\toprule
Category & Keywords \\
\midrule
Anatomy & heart, cardiac, cardiovascular, myocardial, atrial, ventricular, aortic, mitral, coronary, pericardial \\
Conditions & arrhythmia, fibrillation, tachycardia, bradycardia, heart failure, cardiomyopathy, endocarditis, angina, infarction, stenosis, regurgitation \\
Symptoms & chest pain, palpitations, dyspnea on exertion, syncope, orthopnea, edema \\
Procedures & angioplasty, catheterization, ablation, bypass (CABG), stent, pacemaker, defibrillator, echocardiogram, ECG/EKG \\
Medications & anticoagulant, antiarrhythmic, beta-blocker, ACE inhibitor, statin, nitroglycerin, digoxin, warfarin, heparin \\
Risk factors & hypertension, hyperlipidemia, atherosclerosis, cholesterol, blood pressure \\
\bottomrule
\end{tabular}
\suppTableCaption{Keyword dictionary for fallback heart-related classification}{A question is classified as heart-related if it contains at least one keyword from any category. The dictionary is organized by clinical domain and achieves high recall by design.}
\label{tab:keyword_dict}
\end{table}

\begin{figure}[h]
  \centering
  \includegraphics[width=0.85\textwidth]{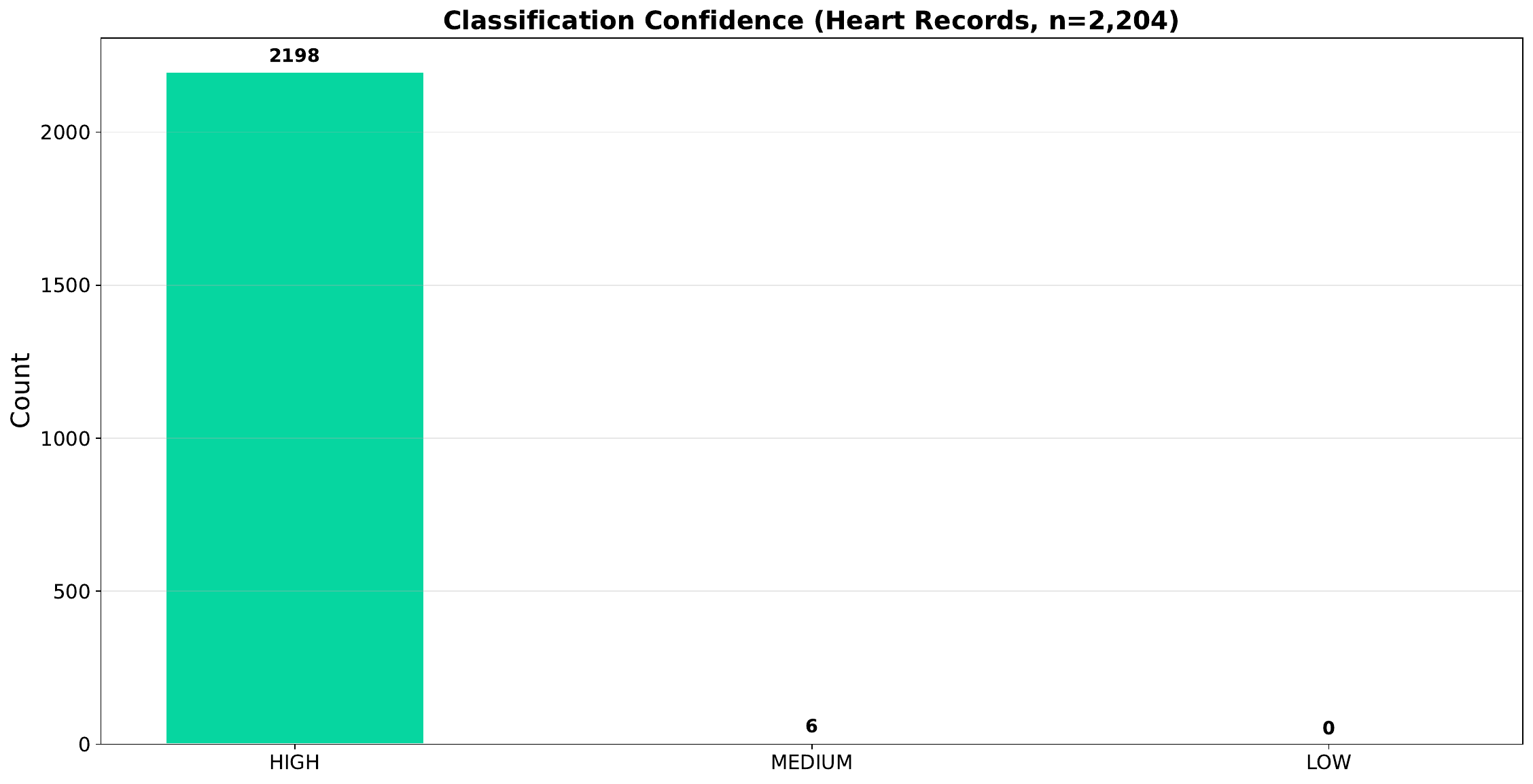}
  \suppFigCaption{Classification confidence distribution}{Distribution of confidence scores from the LLM-based heart-related classifier. High-confidence decisions dominate, indicating that most classification decisions are unambiguous.}
  \label{suppfig:confidence}
\end{figure}

\FloatBarrier

\begin{figure}[h]
  \centering
  \includegraphics[width=0.85\textwidth]{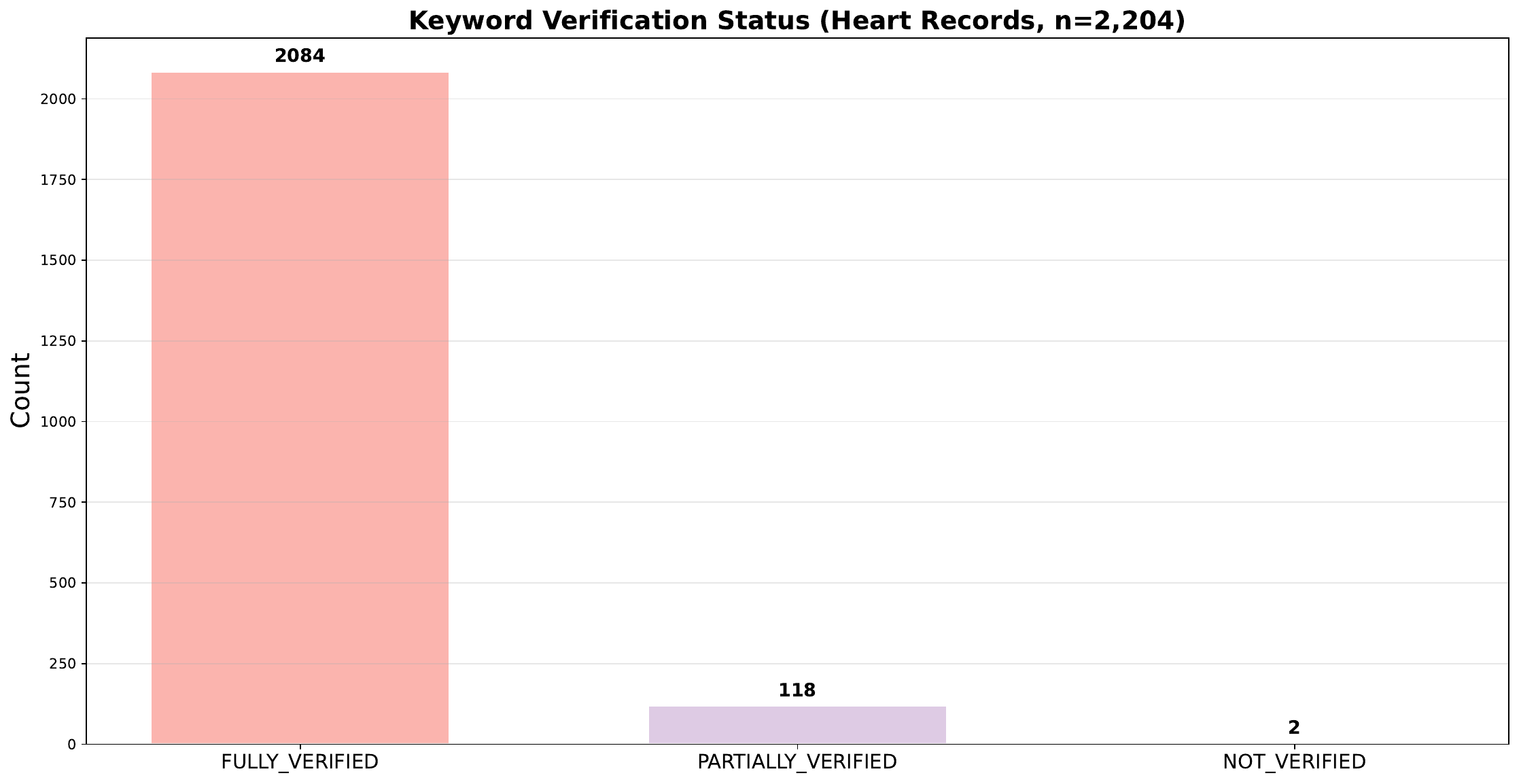}
  \suppFigCaption{Keyword verification status}{Agreement between the LLM-based classifier and the keyword-based fallback classifier, confirming that both identify a consistent set of heart-related questions.}
  \label{suppfig:keyword_verification}
\end{figure}

\FloatBarrier

\section{Supplementary Note 3: Synthetic Reasoning Trace Generation}
\label{supp:sec:synthetic}

When the training examples contain a reference answer but not an explicit reasoning trace, we use a capable language model to generate clinical reasoning consistent with the reference answer. The generated traces are inserted between \texttt{<start\_working\_out>} and \texttt{<end\_working\_out>} tags during SFT transcript construction. When no trace is available, a generic scaffold is used:

\begin{lstlisting}[caption={Generic reasoning scaffold used as fallback.}]
<start_working_out>
Let me analyze this medical question carefully and provide
a thorough response based on clinical knowledge.
<end_working_out>
\end{lstlisting}

Synthetic reasoning is only used during SFT to teach the model the output structure, and the model generates its own reasoning from scratch during GRPO.

\section{Supplementary Note 4: Hyperparameter Tables}
\label{supp:sec:hyperparameters}

The following tables consolidate all hyperparameters referenced in the Methods section of the main paper.

\begin{table}[h]
\centering
\small
\begin{tabular}{l l}
\toprule
Parameter & Value \\
\midrule
Base model & Qwen3-14B-Base (\texttt{unsloth/Qwen3-14B-Base}) \\
Quantization & 4-bit (NF4 via bitsandbytes) \\
Maximum sequence length & 4096 tokens \\
LoRA rank ($r$) & 16 \\
LoRA alpha ($\alpha_{\mathrm{LoRA}}$) & 32 ($= 2r$) \\
LoRA target modules & \texttt{q\_proj}, \texttt{k\_proj}, \texttt{v\_proj}, \texttt{o\_proj}, \\
 & \texttt{gate\_proj}, \texttt{up\_proj}, \texttt{down\_proj} \\
Gradient checkpointing & Enabled (unsloth mode) \\
Random seed & 3407 \\
vLLM GPU memory utilization & 0.8 \\
\bottomrule
\end{tabular}
\suppTableCaption{Model configuration}{Base model, quantization, and LoRA hyperparameters.}
\label{tab:model_config}
\end{table}

\FloatBarrier

\begin{table}[h]
\centering
\small
\begin{tabular}{l l}
\toprule
Parameter & Value \\
\midrule
Per-device batch size & 4 \\
Gradient accumulation steps & 4 \\
Effective batch size & 16 \\
Learning rate & $2 \times 10^{-4}$ \\
Optimizer & AdamW (8-bit) \\
Weight decay & $1 \times 10^{-3}$ \\
LR scheduler & Linear \\
Warmup steps & 5 \\
Number of epochs & 2 \\
Max training steps & 500 \\
Instruction masking & System + user tokens (label $= -100$) \\
\bottomrule
\end{tabular}
\suppTableCaption{SFT training hyperparameters}{Instruction masking trains only on assistant response tokens.}
\label{tab:sft_hyperparams}
\end{table}

\FloatBarrier

\begin{table}[h]
\centering
\small
\begin{tabular}{l l}
\toprule
Parameter & Value \\
\midrule
Per-device batch size & 1 \\
Gradient accumulation steps & 1 \\
Number of generations per prompt ($G$) & 6 \\
Learning rate & $5 \times 10^{-6}$ \\
Optimizer & AdamW (8-bit) \\
Weight decay & $1 \times 10^{-2}$ \\
LR scheduler & Linear \\
Warmup ratio & 0.1 \\
Max training steps & 1000 \\
Max prompt length & 90th percentile of training prompts \\
Max completion length & 1024 tokens \\
Temperature (generation) & 1.0 \\
\texttt{min\_p} / \texttt{top\_p} / \texttt{top\_k} & 0.1 / 1.0 / $-1$ (disabled) \\
Minimum criteria count & $> 3$ \\
\bottomrule
\end{tabular}
\suppTableCaption{GRPO training hyperparameters}{GRPO samples $G=6$ completions per prompt and filters out prompts with $\leq 3$ rubric criteria.}
\label{tab:grpo_hyperparams}
\end{table}

\FloatBarrier

\begin{table}[h]
\centering
\small
\begin{tabular}{l c c l}
\toprule
Parameter & Hybrid & Complexity & Description \\
\midrule
$r_{\mathrm{base}}$ & 20 & 20 & Base reward scale \\
$B$ (base component) & 15 & --- & Continuous base portion \\
$P$ (perfection bonus) & 5 & --- & Bonus for perfect satisfaction \\
$\alpha$ (exponent) & --- & 1.2 & Score curvature \\
$\beta$ (complexity coeff.) & --- & 0.2 & Logarithmic complexity bonus \\
$\lambda$ (negative penalty) & 0.3 & 0.5 & Penalty coefficient for negative criteria \\
$n_{\mathrm{max}}$ & --- & 25 & Maximum rubric count normalization \\
Reward range & $[0, 20]$ & $[0, \sim\!25]$ & Output range \\
\bottomrule
\end{tabular}
\suppTableCaption{Reward function hyperparameters}{Comparison of the Hybrid and Complexity-aware reward variants.}
\label{tab:reward_hyperparams}
\end{table}

\FloatBarrier

\begin{table}[h]
\centering
\small
\begin{tabular}{l l}
\toprule
Parameter & Value \\
\midrule
Judge model & GPT-OSS-120B (\texttt{openai/gpt-oss-120b}) \\
API provider & Groq \\
Temperature & 0.1 \\
Response format & JSON object (structured output) \\
Max completion tokens & 1024 (2048 for criteria $> 1000$ chars) \\
Output validation & Pydantic schema (\texttt{present}, \texttt{justification}) \\
Retry attempts per criterion & 3 \\
\bottomrule
\end{tabular}
\suppTableCaption{Judge model configuration}{Settings for the LLM-as-judge used during GRPO training and evaluation.}
\label{tab:judge_config}
\end{table}

\FloatBarrier

\begin{table}[h]
\centering
\small
\begin{tabular}{l l}
\toprule
Parameter & Value \\
\midrule
Evaluation set & HealthBench (heart-related subset, held out) \\
Number of evaluation samples & 500 \\
Sample random seed & 42 \\
Generation temperature (local) & 0.0 \\
Generation temperature (Groq) & 0.0 \\
Max generation tokens & 1024 \\
\texttt{top\_k} (local) & 1 \\
Rubric scoring method & LLM-as-judge criterion-level evaluation \\
\bottomrule
\end{tabular}
\suppTableCaption{Evaluation configuration}{Deterministic decoding (temperature~0) ensures reproducible evaluation outputs.}
\label{tab:eval_config}
\end{table}

\FloatBarrier

\section{Supplementary Note 5: Evaluated Models}
\label{supp:sec:models}

\begin{table}[h]
\centering
\small
\begin{tabular}{l l l}
\toprule
Display Name & Model Identifier & Interface \\
\midrule
Kimi-K2 & \texttt{moonshotai/kimi-k2-instruct-0905} & Groq API \\
Llama-4-Scout-17B & \texttt{meta-llama/llama-4-scout-17b-16e-instruct} & Groq API \\
Llama-4-Maverick-17B & \texttt{meta-llama/llama-4-maverick-17b-128e-instruct} & Groq API \\
Llama-3.3-70B & \texttt{llama-3.3-70b-versatile} & Groq API \\
GPT-OSS-120B & \texttt{openai/gpt-oss-120b} & Groq API \\
\midrule
MedGemma-1.5-4B & \texttt{MedAIBase/MedGemma1.5:4b} & Ollama \\
MedGemma-4B & \texttt{alibayram/medgemma:4b} & Ollama \\
MedGemma-27B & \texttt{alibayram/medgemma:27b} & Ollama \\
Qwen3-14B (Ollama) & \texttt{qwen3:14b} & Ollama \\
Gemma3-12B & \texttt{gemma3:12b} & Ollama \\
Phi4-14B & \texttt{phi4:14b} & Ollama \\
\midrule
Qwen3-14B Base & \texttt{unsloth/Qwen3-14B-Base} + SFT LoRA & vLLM \\
GRPO (HYBRID) & + Hybrid reward LoRA & vLLM \\
GRPO (COMPLEXITY) & + Complexity reward LoRA & vLLM \\
GRPO (NEURAL) & + Neural reward LoRA & vLLM \\
GRPO (RaR-EXPLICIT) & + RaR-Explicit reward LoRA & vLLM \\
GRPO (RaR-IMPLICIT) & + RaR-Implicit reward LoRA & vLLM \\
\bottomrule
\end{tabular}
\suppTableCaption{Complete model inventory}{The study considered 17 model configurations on the held-out HealthBench heart subset ($n=500$). The primary performance tables and figures summarize the 15 configurations with final benchmark metrics, while Qwen3-14B (Ollama) and GRPO (NEURAL) are retained in ablation and response time analyses. Local LoRA models are stacked on the shared Qwen3-14B-Base + SFT checkpoint, differing only in the GRPO reward function used for post-training.}
\label{tab:all_models}
\end{table}

\FloatBarrier

\section{Supplementary Note 6: Reward Function Ablation}
\label{supp:sec:reward_ablation}

This section extends the reward analysis in the ablation section of the main paper with distributional statistics from the full set of 6{,}000 GRPO training completions (1{,}000 optimization steps $\times$ $G=6$ generations). We evaluated eight candidate reward functions on the same judge outputs to isolate the effect of reward shaping.

\subsection{Distributional comparison}

\begin{table}[h]
\centering
\small
\begin{tabular}{l r r r r r}
\toprule
Reward Function & Mean & Std & Non-zero (\%) & Perfect & Corr.\ w/~$s_{\mathrm{norm}}$ \\
\midrule
Binary (pass/fail) & 0.07 & 1.18 & 0.35 & 21 & --- \\
Linear & 3.73 & 4.22 & 62.2 & 22 & 1.000 \\
Threshold-linear & 0.57 & 2.25 & 10.0 & 22 & --- \\
Exponential & 1.59 & 2.79 & 62.2 & 22 & --- \\
Negative-penalty & 1.84 & 3.42 & 34.0 & 21 & --- \\
Hybrid (ours) & 2.81 & 3.25 & 62.2 & 21 & 0.975 \\
Complexity-aware (ours) & 5.55 & 6.26 & 62.2 & 194 & 0.990 \\
Neural & 5.21 & 4.14 & 100.0 & 0 & 0.953 \\
\bottomrule
\end{tabular}
\suppTableCaption{Reward function distributional statistics over 6{,}000 GRPO completions}{``Non-zero'' is the percentage with reward~$>0$. ``Perfect'' counts completions achieving the maximum score. ``Corr.'' is Pearson correlation with the underlying normalized rubric score. The binary reward places 99.65\% of mass at zero, making it unsuitable for GRPO's group normalization. The Hybrid and Complexity-aware rewards provide intermediate non-zero rates with meaningful spread.}
\label{tab:reward_comparison}
\end{table}

\FloatBarrier

\subsection{Binary reward sparsity}

Under the binary (pass/fail) reward, only 21 of 6{,}000 completions (0.35\%) received non-zero reward. All 21 achieved the maximum score of~20. The remaining 5{,}979 completions received zero reward regardless of partial quality---including completions with normalized rubric scores of 0.92, 0.89, and 0.88 that satisfied the vast majority of positive criteria. When all $G=6$ completions in a GRPO group receive zero reward, the group-normalized advantage is exactly zero and the policy update provides no learning signal.

In contrast, the Hybrid and Complexity-aware rewards achieve 62.2\% non-zero rates with standard deviations of 3.25 and 6.26, creating within-group variance on most prompts and enabling consistent policy improvement.

\subsection{Complexity bonus effect}

The Complexity-aware reward has 194 perfect-score completions (vs.\ 21--22 for other variants) because the logarithmic complexity bonus allows high-performing completions on complex prompts to exceed the base reward scale. The bonus is bounded by $\beta \cdot \log(1+n_c)/\log(1+n_{\max}) \leq 0.2$, so the maximum reward is approximately $20 \times 1.2 \approx 24$ for the most complex prompts. This adds further separation between high quality completions without skewing the overall scale.

\section{Supplementary Note 7: Pairwise McNemar Significance Tests}
\label{supp:sec:mcnemar}

This section provides the full pairwise McNemar test results referenced in the ablation section and Fig.~5 of the main paper. McNemar's test assesses whether two models make significantly different errors on the same 500-sample evaluation set. The test statistic is $\chi^2 = (n_{10} - n_{01})^2 / (n_{10} + n_{01})$, where $n_{10}$ and $n_{01}$ are discordant pairs.

\subsection{GRPO variants vs.\ baselines}

\begin{table}[h]
\centering
\small
\begin{tabular}{l l r r l}
\toprule
Model A & Model B & $\chi^2$ & $p$-value & Sig. \\
\midrule
Base & GRPO (HYBRID) & 32.06 & $1.5 \times 10^{-8}$ & ** \\
Base & GRPO (COMPLEXITY) & 39.03 & $4.2 \times 10^{-10}$ & ** \\
Base & GRPO (NEURAL) & 0.77 & 0.382 & n.s. \\
Base & GRPO (RaR-EXPLICIT) & 2.75 & 0.097 & n.s. \\
Base & GRPO (RaR-IMPLICIT) & 5.70 & 0.017 & * \\
\midrule
GRPO (HYBRID) & GRPO (COMPLEXITY) & 0.01 & 0.928 & n.s. \\
GRPO (HYBRID) & GRPO (RaR-EXPLICIT) & 19.08 & $1.3 \times 10^{-5}$ & ** \\
GRPO (HYBRID) & GRPO (RaR-IMPLICIT) & 13.26 & $2.7 \times 10^{-4}$ & ** \\
GRPO (COMPLEXITY) & GRPO (RaR-EXPLICIT) & 20.33 & $6.5 \times 10^{-6}$ & ** \\
GRPO (COMPLEXITY) & GRPO (RaR-IMPLICIT) & 15.24 & $9.5 \times 10^{-5}$ & ** \\
\bottomrule
\end{tabular}
\suppTableCaption{McNemar tests: GRPO reward variants}{Significance levels: ** $p < 0.01$, * $p < 0.05$, n.s.\ not significant. The Hybrid and Complexity-aware rewards produce statistically significant improvements over the base model and over both RaR aggregation strategies. The two variance-aware rewards are not significantly different from each other.}
\label{tab:mcnemar_grpo}
\end{table}

\FloatBarrier

\subsection{GRPO variants vs.\ external models}

\begin{table}[h]
\centering
\small
\begin{tabular}{l l r r l}
\toprule
Model A & Model B & $\chi^2$ & $p$-value & Sig. \\
\midrule
GRPO (HYBRID) & GPT-OSS-120B & 0.11 & 0.746 & n.s. \\
GRPO (COMPLEXITY) & GPT-OSS-120B & 0.03 & 0.860 & n.s. \\
GRPO (HYBRID) & Kimi-K2 & 8.17 & 0.004 & ** \\
GRPO (COMPLEXITY) & Kimi-K2 & 6.98 & 0.008 & ** \\
\midrule
GRPO (HYBRID) & Llama-4-Scout-17B & 7.26 & 0.007 & ** \\
GRPO (HYBRID) & Llama-4-Maverick-17B & 18.01 & $2.2 \times 10^{-5}$ & ** \\
GRPO (HYBRID) & Llama-3.3-70B & 15.59 & $7.9 \times 10^{-5}$ & ** \\
GRPO (HYBRID) & MedGemma-27B & 3.72 & 0.054 & n.s. \\
GRPO (COMPLEXITY) & MedGemma-27B & 4.86 & 0.027 & * \\
\midrule
GRPO (HYBRID) & Gemma3-12B & 5.28 & 0.022 & * \\
GRPO (HYBRID) & Phi4-14B & 5.98 & 0.015 & * \\
GRPO (COMPLEXITY) & Gemma3-12B & 6.09 & 0.014 & * \\
GRPO (COMPLEXITY) & Phi4-14B & 6.68 & 0.010 & ** \\
\bottomrule
\end{tabular}
\suppTableCaption{McNemar tests: GRPO variants vs.\ external models}{Neither variance-aware GRPO variant is significantly different from GPT-OSS-120B ($p > 0.7$), indicating statistically equivalent performance despite the GRPO models being ${\sim}9\times$ smaller. Both are significantly outperformed by Kimi-K2 only.}
\label{tab:mcnemar_external}
\end{table}

\FloatBarrier

\subsection{Confidence interval methodology}

The 95\% confidence intervals reported in Table~2 of the main paper use the normal approximation for a binomial proportion:
\begin{equation}
\hat{p} \pm 1.96 \sqrt{\frac{\hat{p}(1-\hat{p})}{n}}, \qquad n = 500.
\end{equation}
This approximation is appropriate because $n\hat{p}$ and $n(1-\hat{p})$ both exceed~5 for all models. For the 17 model configurations considered in the broader evaluation, there are $\binom{17}{2} = 136$ pairwise tests. The key comparisons of interest (GRPO variants vs.\ base model and vs.\ frontier models) achieve $p < 0.01$ and remain significant under Bonferroni correction.

The clinician review is reported separately in Section~1.4 and Table~4 of the main paper. Its response level proportions use Wilson score 95\% confidence intervals for the 56 reviewed responses, as specified in the main Methods section.

\section{Supplementary Note 8: Response Time Analysis}
\label{supp:sec:response_times}

\begin{table}[h]
\centering
\small
\begin{tabular}{l l r r r r}
\toprule
Model & Interface & Mean (s) & Median (s) & P95 (s) & Total (s) \\
\midrule
Qwen3-14B Base & vLLM & 5.36 & 5.18 & 11.23 & 2682 \\
GRPO (HYBRID) & vLLM & 9.72 & 9.44 & 12.96 & 4862 \\
GRPO (COMPLEXITY) & vLLM & 9.90 & 9.85 & 12.95 & 4949 \\
GRPO (NEURAL) & vLLM & 6.17 & 5.96 & 12.91 & 3085 \\
GRPO (RaR-EXPLICIT) & vLLM & 6.15 & 6.31 & 12.93 & 3077 \\
GRPO (RaR-IMPLICIT) & vLLM & 6.15 & 6.36 & 12.91 & 3074 \\
\midrule
Kimi-K2 & Groq & 5.10 & 4.77 & 9.98 & 2548 \\
Llama-4-Scout-17B & Groq & 1.48 & 1.51 & 1.97 & 742 \\
Llama-4-Maverick-17B & Groq & 1.16 & 1.17 & 1.67 & 578 \\
Llama-3.3-70B & Groq & 2.08 & 2.06 & 2.95 & 1038 \\
GPT-OSS-120B & Groq & 2.17 & 2.18 & 2.22 & 1086 \\
\midrule
MedGemma-27B & Ollama & 20.26 & 20.74 & 24.96 & 10130 \\
MedGemma-4B & Ollama & 6.46 & 6.48 & 10.23 & 3228 \\
MedGemma-1.5-4B & Ollama & 8.71 & 8.67 & 13.20 & 4354 \\
Gemma3-12B & Ollama & 7.74 & 8.07 & 8.16 & 3870 \\
Phi4-14B & Ollama & 3.70 & 3.71 & 4.82 & 1850 \\
Qwen3-14B (Ollama) & Ollama & 7.52 & 7.74 & 7.83 & 3760 \\
\bottomrule
\end{tabular}
\suppTableCaption{Per-model response time statistics over 500 evaluation samples}{The GRPO (HYBRID) and GRPO (COMPLEXITY) models produce longer responses (mean~${\sim}10$\,s) than the base model (${\sim}5$\,s), consistent with more extensive reasoning traces. Groq API models show lower latency due to dedicated inference hardware; Ollama models run locally on the same GPU. Response times include both generation and any network overhead.}
\label{tab:response_times}
\end{table}

\FloatBarrier

The GRPO-trained models with variance-aware rewards produce substantially longer responses than the base model (mean 9.7--9.9\,s vs.\ 5.4\,s), reflecting the extended reasoning traces learned during RL. The RaR-based and Neural reward variants do not show this increase (${\sim}6.2$\,s), which could partially explain their lower rubric satisfaction: shorter responses are less likely to cover all criteria in complex prompts.

\section{Supplementary Note 9: GRPO Training Dynamics}
\label{supp:sec:training_dynamics}

\begin{figure}[h]
  \centering
  \includegraphics[width=\textwidth]{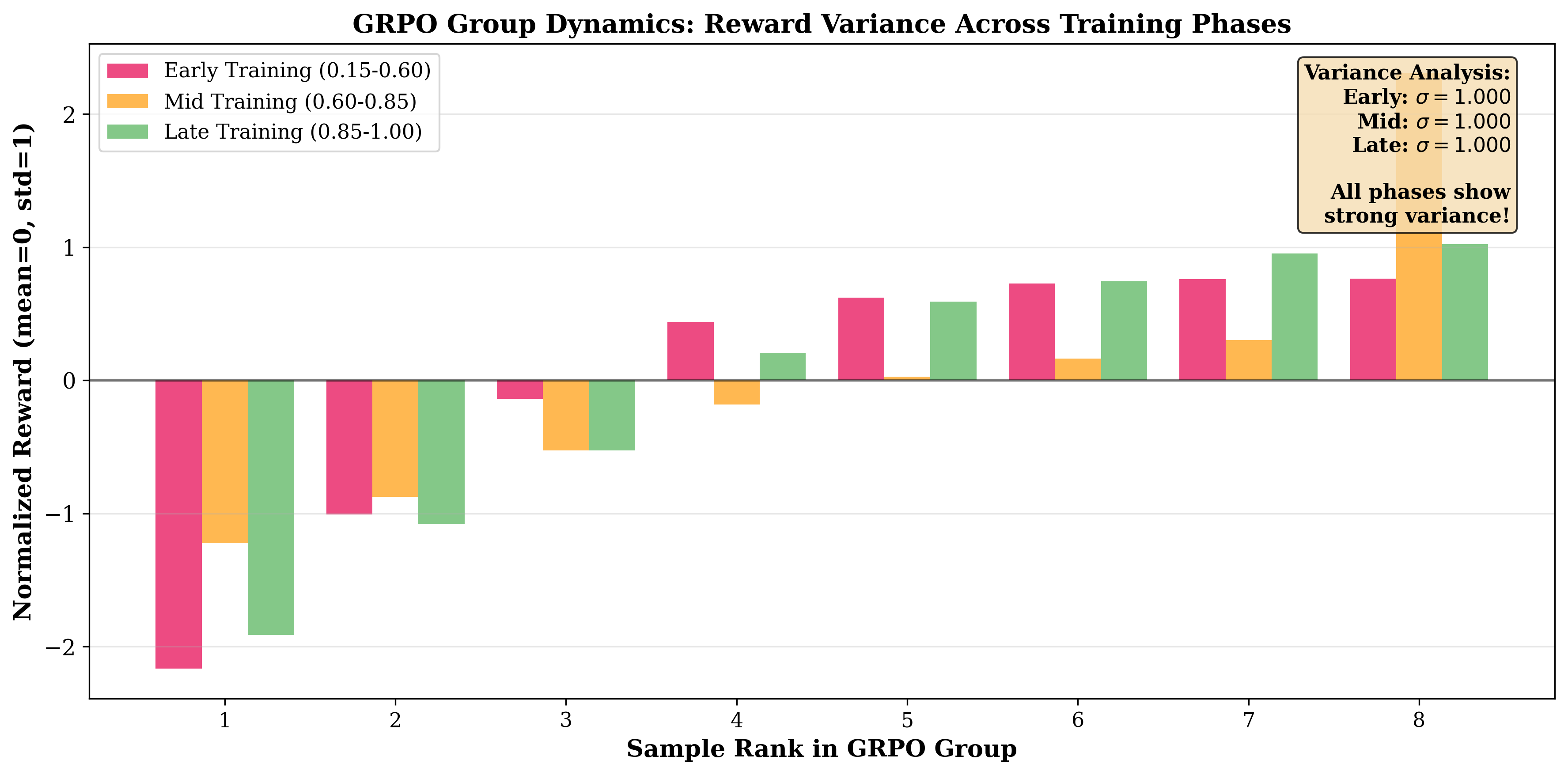}
  \suppFigCaption{GRPO training dynamics}{Evolution of training metrics over 1{,}000 GRPO optimization steps showing reward progression and convergence behavior for the Hybrid and Complexity-aware reward variants.}
  \label{suppfig:grpo_dynamics}
\end{figure}

\FloatBarrier

The training dynamics exhibit three phases: (1)~initial exploration (steps 0--100) with high reward variance as the SFT-initialized policy begins sampling diverse completions; (2)~rapid improvement (steps 100--500) with accelerating mean reward and decreasing intra-group variance; and (3)~plateau (steps 500--1000) where remaining gains come from harder prompts with larger rubric sets. The variance-aware rewards achieved close to final performance before 1{,}000 steps, indicating that shorter training runs may be sufficient for similar tasks in the future.

\section{Supplementary Note 10: Supplementary Dataset Figures}
\label{supp:sec:dataset_stats}

These plots give detailed statistical views of the heart-specific subset of the RaR-Medicine training data. These complement the overall view of the training data in the main paper (Figure 3).

\begin{figure}[h]
  \centering
  \includegraphics[width=0.85\textwidth]{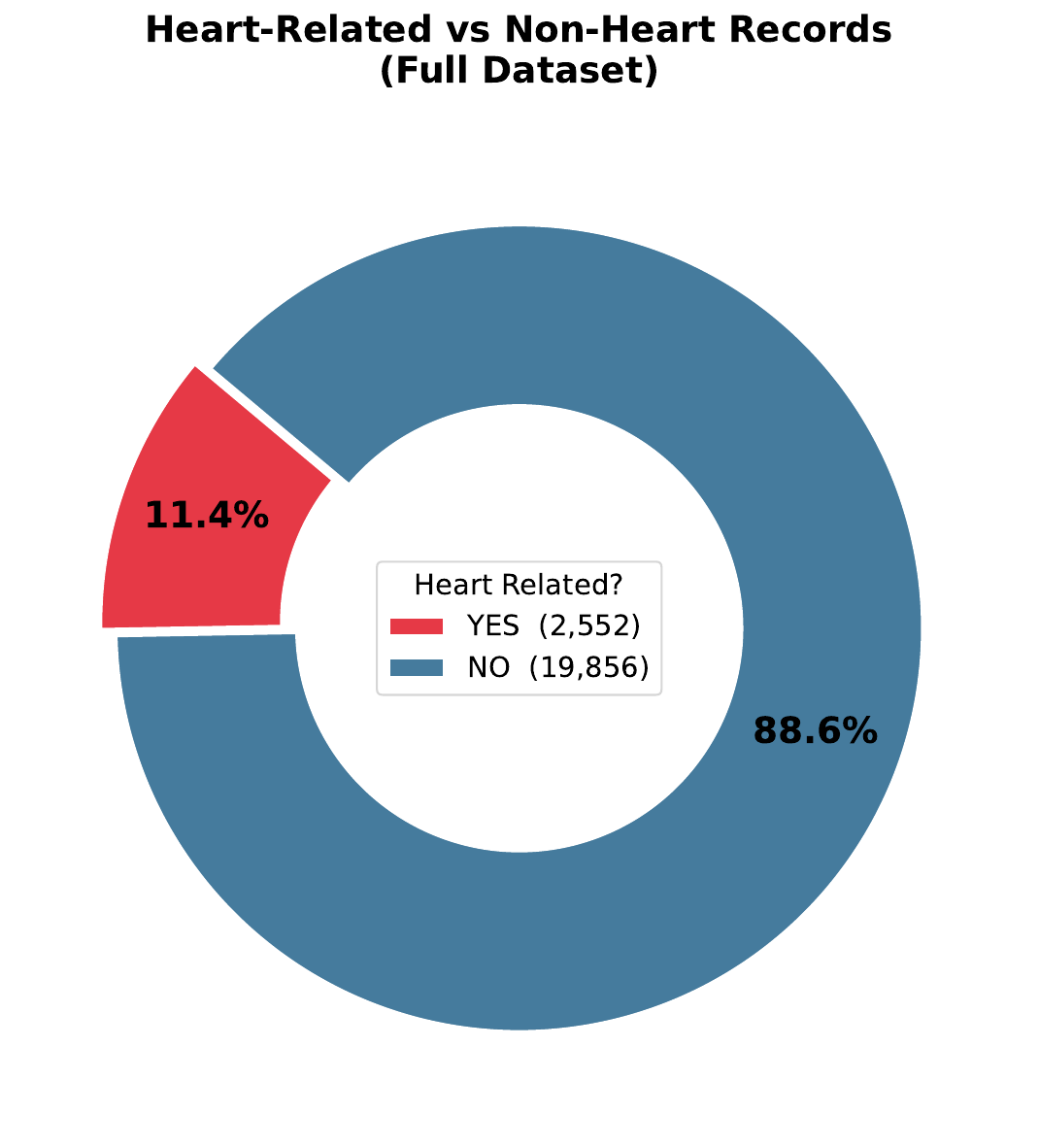}
  \suppFigCaption{Heart-related vs.\ non-heart-related sample distribution}{Proportion of samples classified as heart-related in the filtered RaR-Medicine dataset.}
  \label{suppfig:heart_donut}
\end{figure}

\FloatBarrier

\begin{figure}[h]
  \centering
  \includegraphics[width=\textwidth]{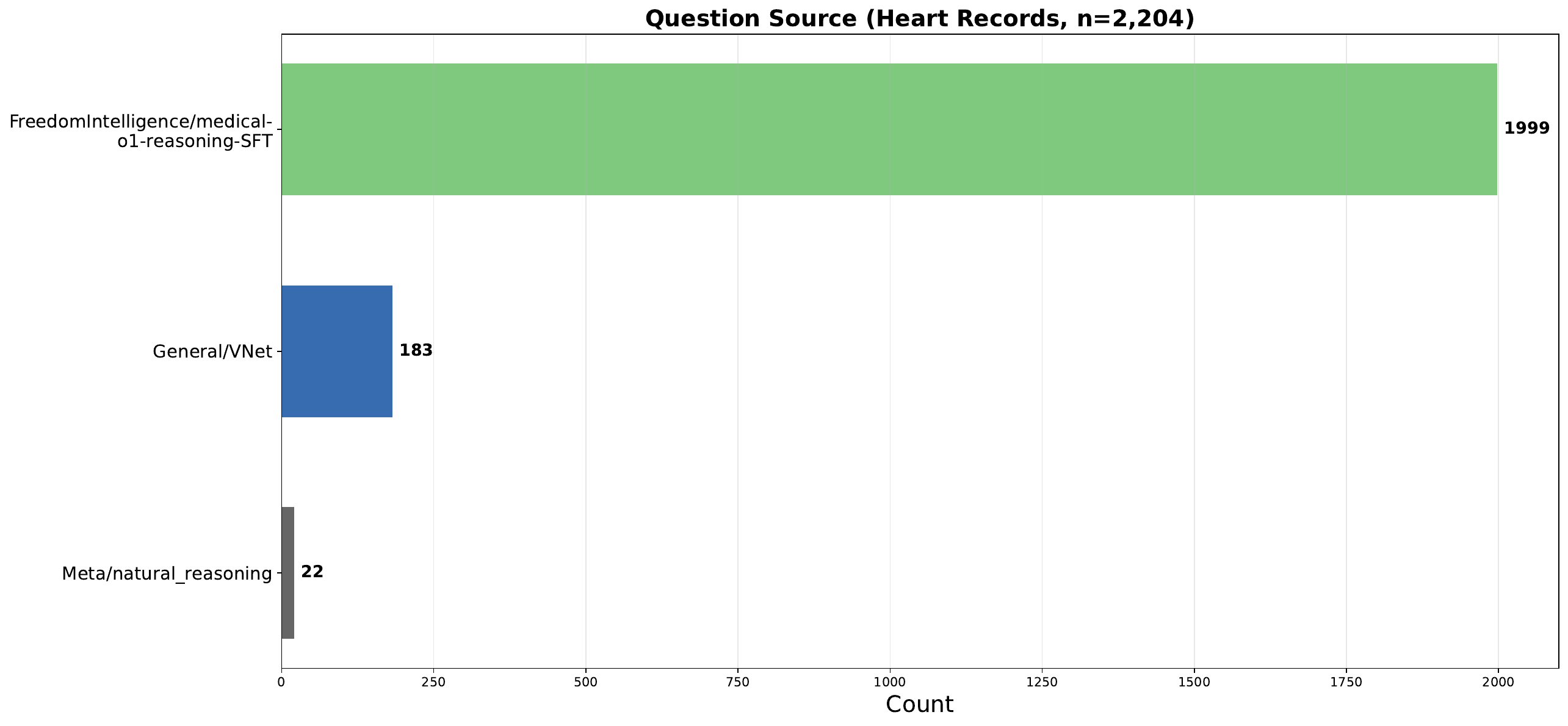}
  \suppFigCaption{Question source distribution}{Horizontal bar chart of question sources in the heart-related RaR-Medicine subset.}
  \label{suppfig:question_source}
\end{figure}

\FloatBarrier

\begin{figure}[h]
  \centering
  \includegraphics[width=\textwidth]{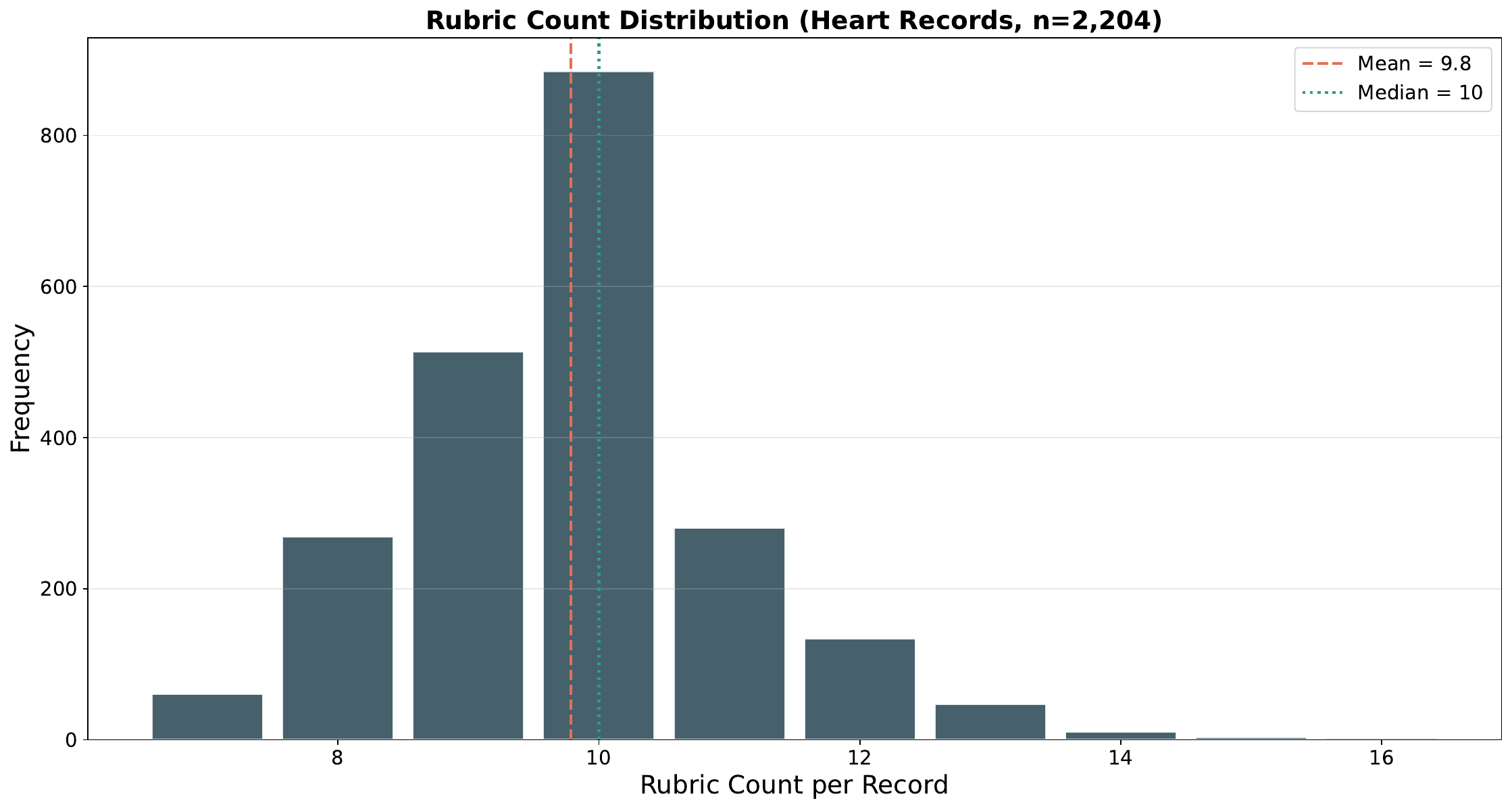}
  \suppFigCaption{Rubric count distribution}{Number of rubric criteria per prompt. The vertical dashed line marks the GRPO filtering threshold ($> 3$ criteria).}
  \label{suppfig:rubric_count}
\end{figure}

\FloatBarrier

\begin{figure}[h]
  \centering
  \includegraphics[width=\textwidth]{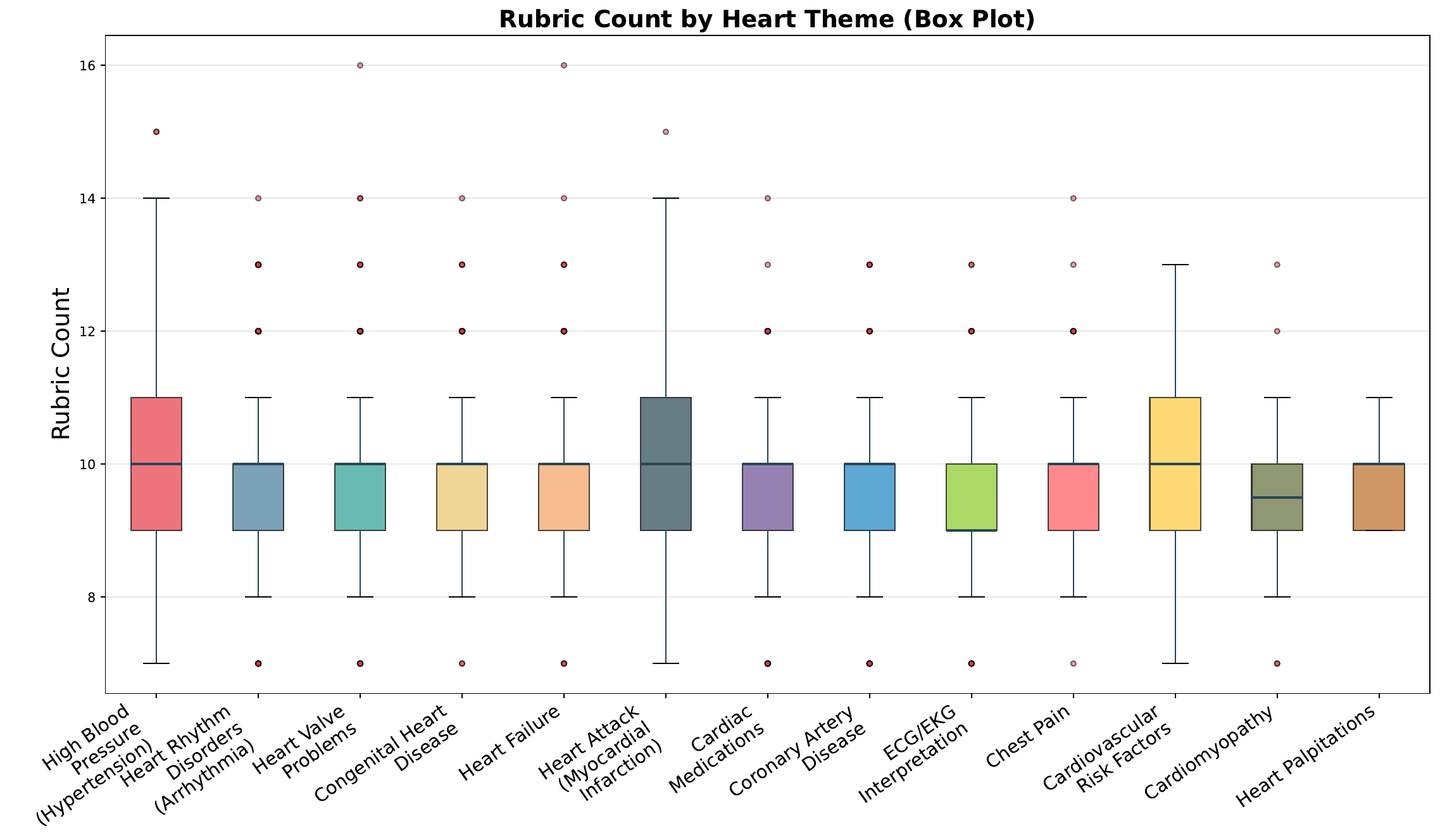}
  \suppFigCaption{Rubric count by theme}{Distribution of rubric criteria counts across heart-related themes.}
  \label{suppfig:rubric_by_theme}
\end{figure}

\FloatBarrier

\begin{figure}[h]
  \centering
  \includegraphics[width=\textwidth]{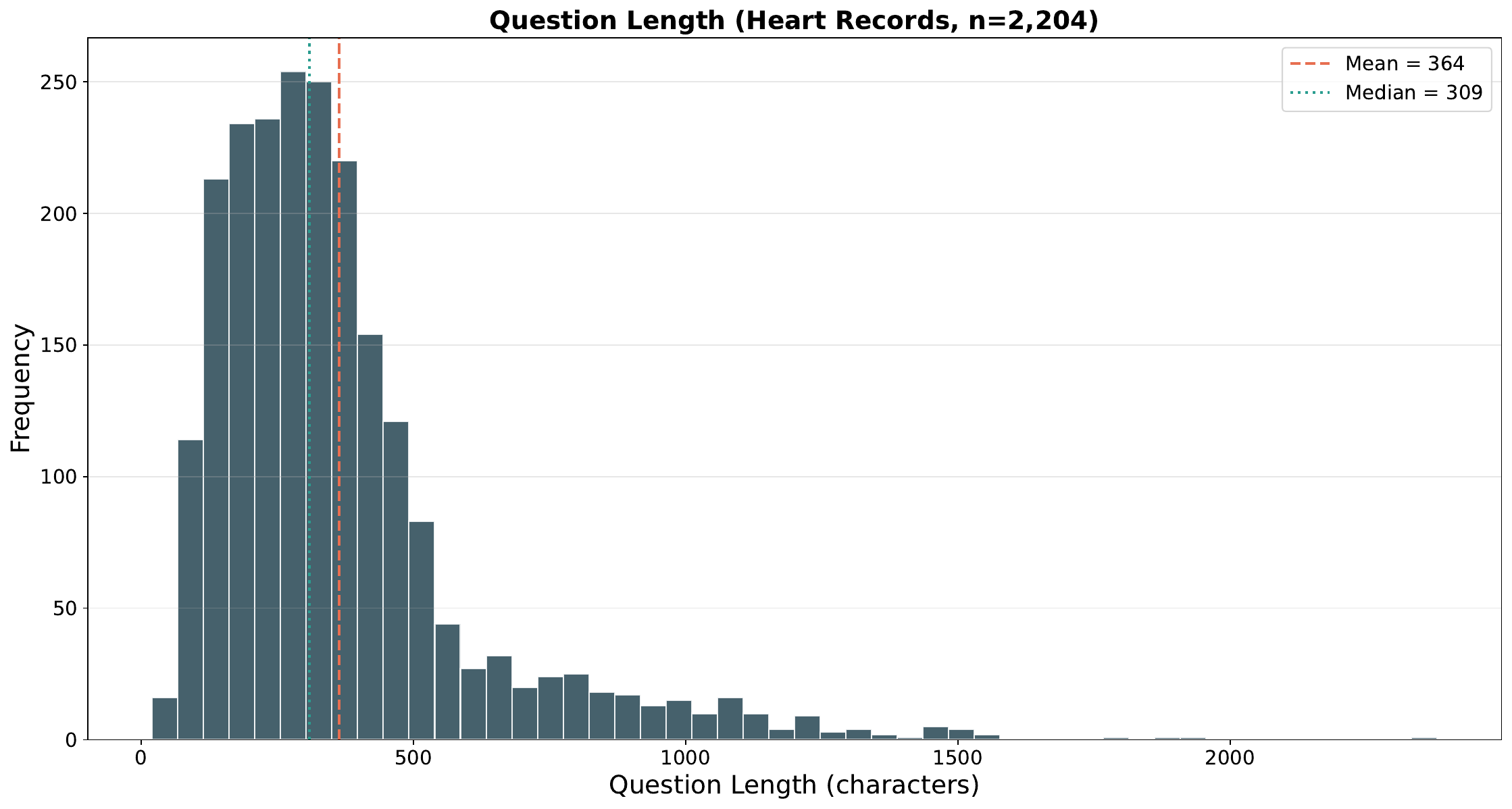}
  \suppFigCaption{Question length distribution}{Histogram of question lengths (characters) in the heart-related RaR-Medicine subset.}
  \label{suppfig:question_length}
\end{figure}

\FloatBarrier

\begin{figure}[h]
  \centering
  \includegraphics[width=\textwidth]{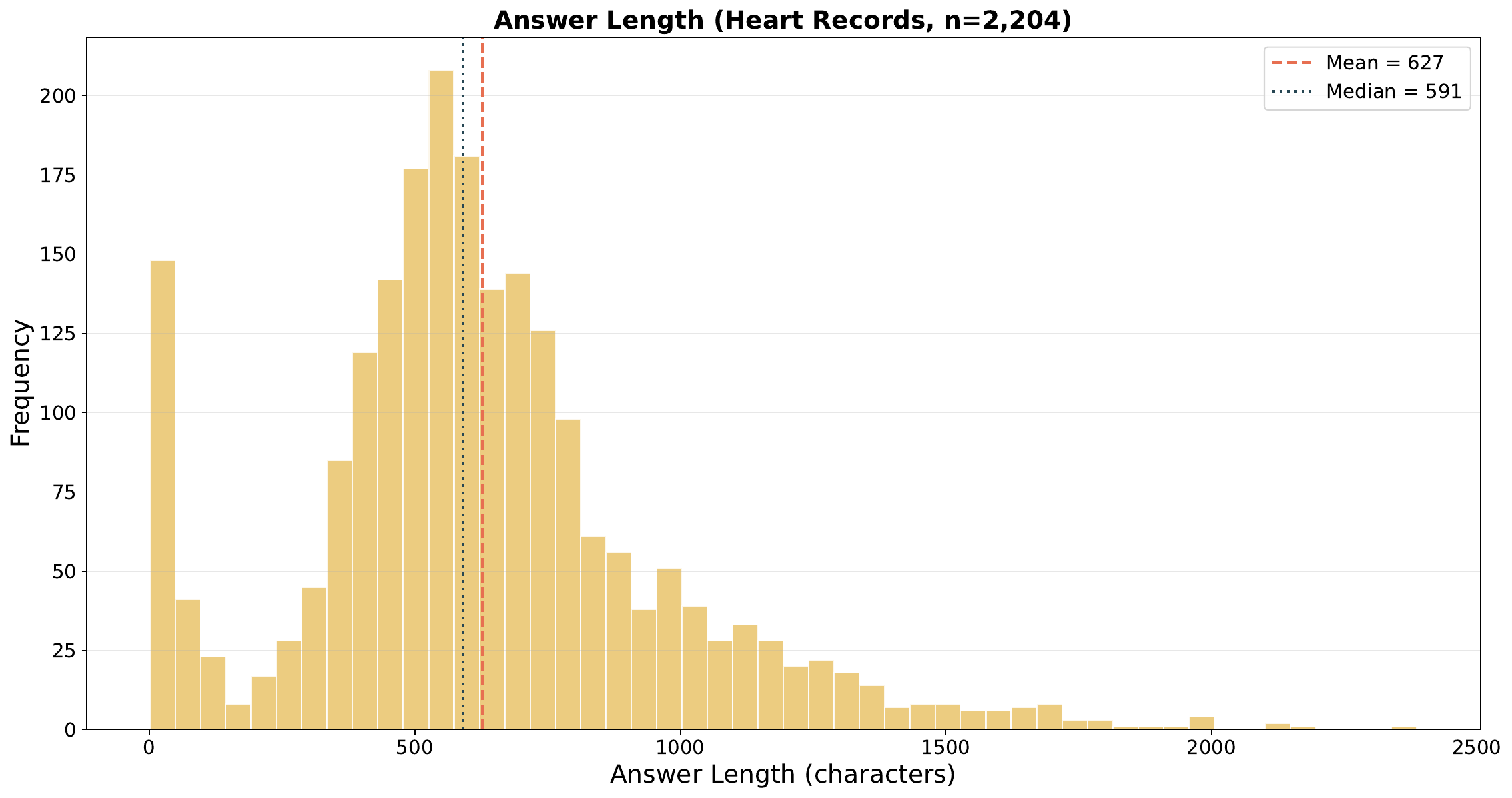}
  \suppFigCaption{Reference answer length distribution}{Histogram of reference answer lengths (characters).}
  \label{suppfig:answer_length}
\end{figure}

\FloatBarrier

\begin{figure}[h]
  \centering
  \includegraphics[width=\textwidth]{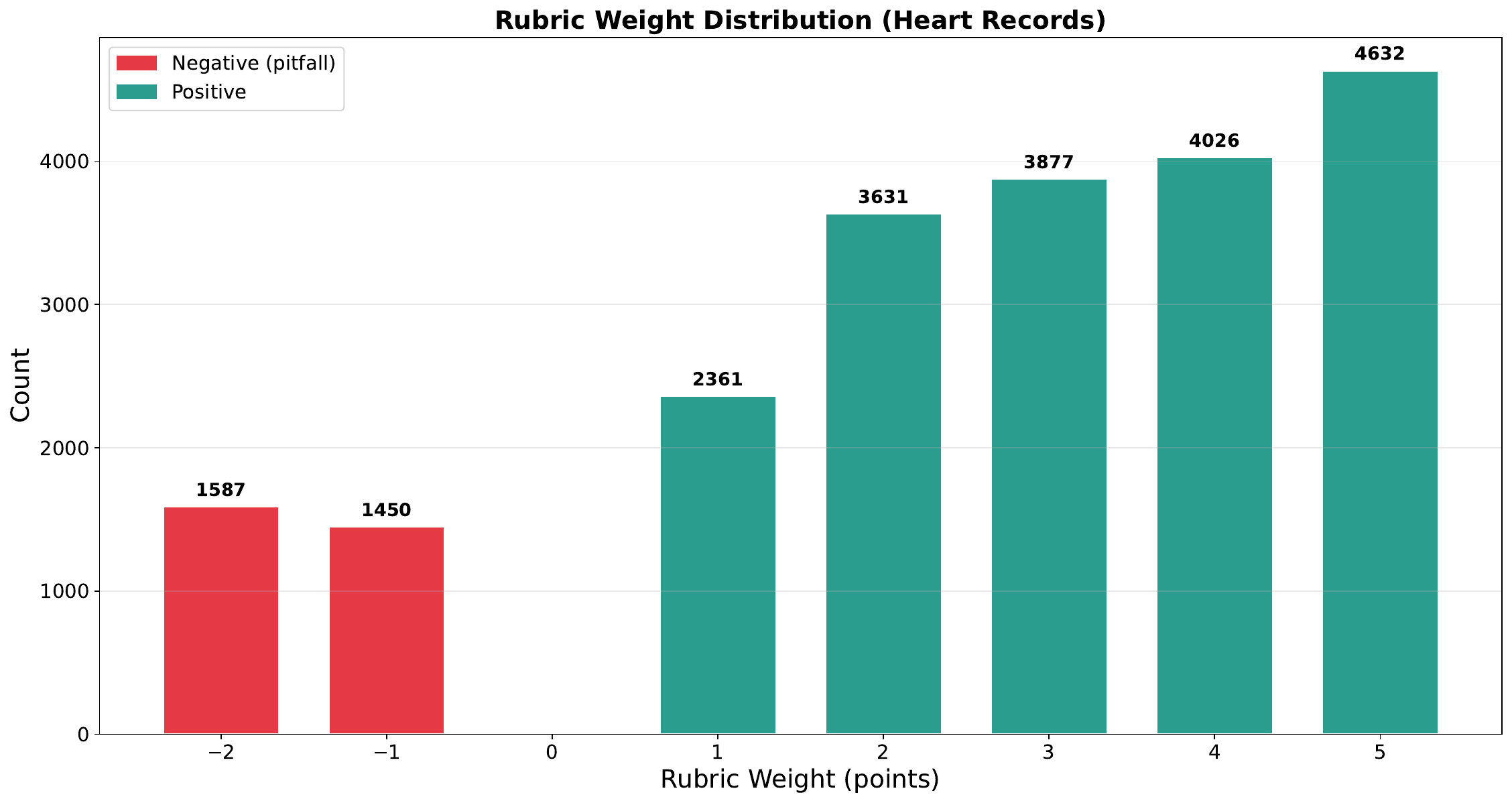}
  \suppFigCaption{Rubric weight distribution}{Distribution of rubric point values. Positive weights reward desirable behaviors; negative weights penalize unsafe content.}
  \label{suppfig:rubric_weights}
\end{figure}

\FloatBarrier

\begin{figure}[h]
  \centering
  \includegraphics[width=\textwidth]{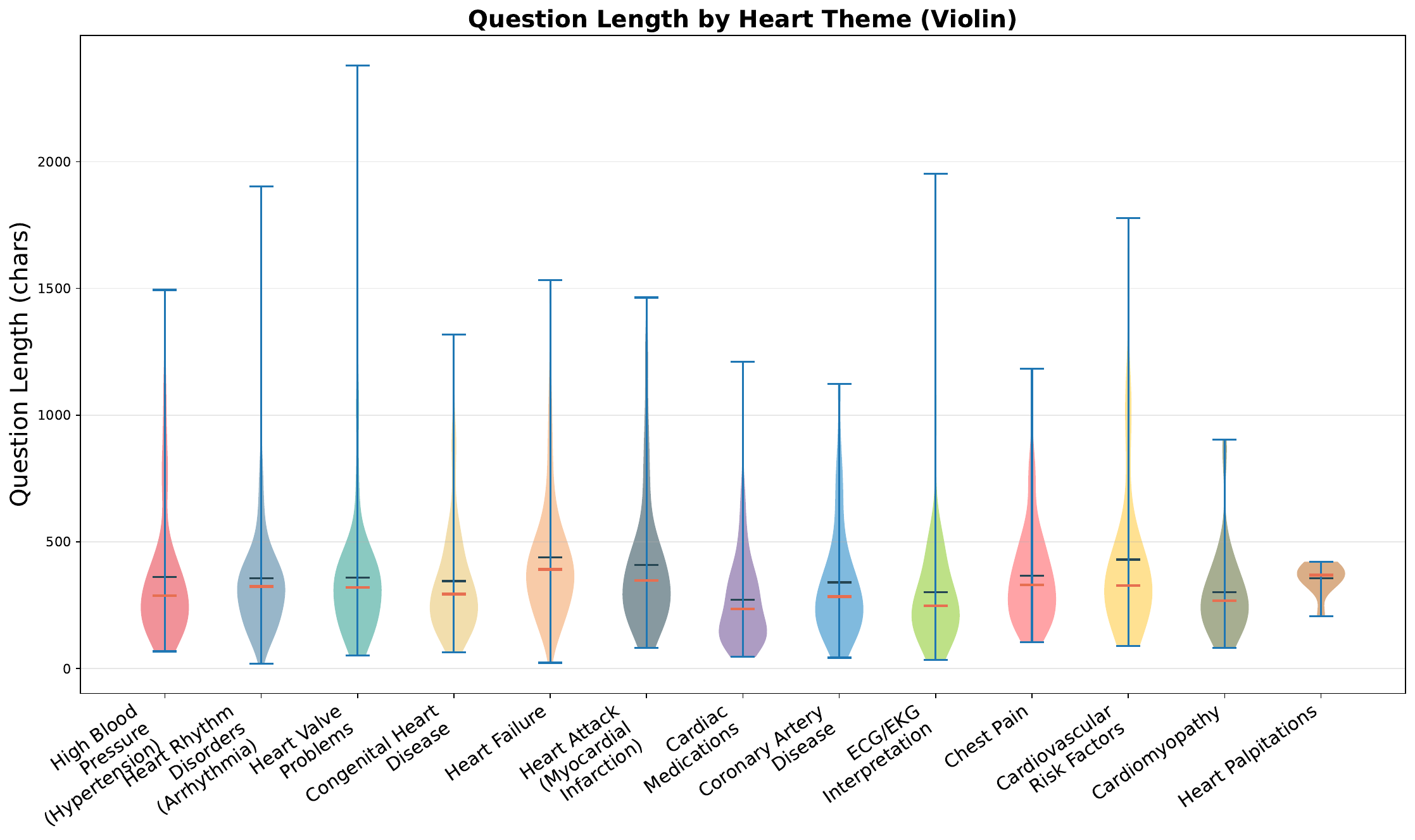}
  \suppFigCaption{Question length by theme (violin plot)}{Distribution of question lengths across heart-related themes. High variability in some themes motivates the 90th-percentile length filter for GRPO.}
  \label{suppfig:question_length_violin}
\end{figure}

\FloatBarrier

\begin{figure}[h]
  \centering
  \includegraphics[width=\textwidth]{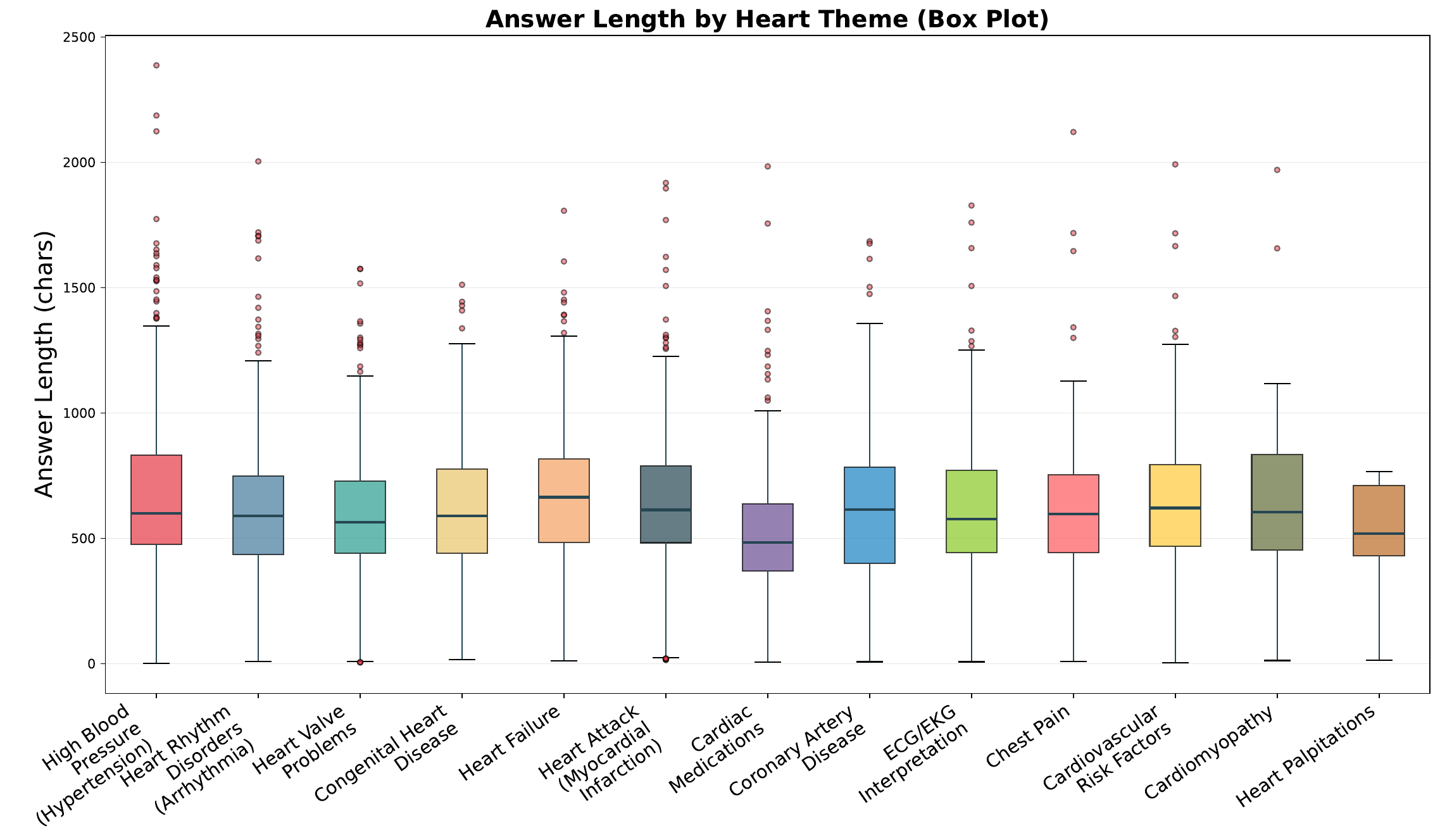}
  \suppFigCaption{Answer length by theme}{Reference answer lengths stratified by heart-related theme.}
  \label{suppfig:answer_by_theme}
\end{figure}

\FloatBarrier

\begin{figure}[h]
  \centering
  \includegraphics[width=\textwidth]{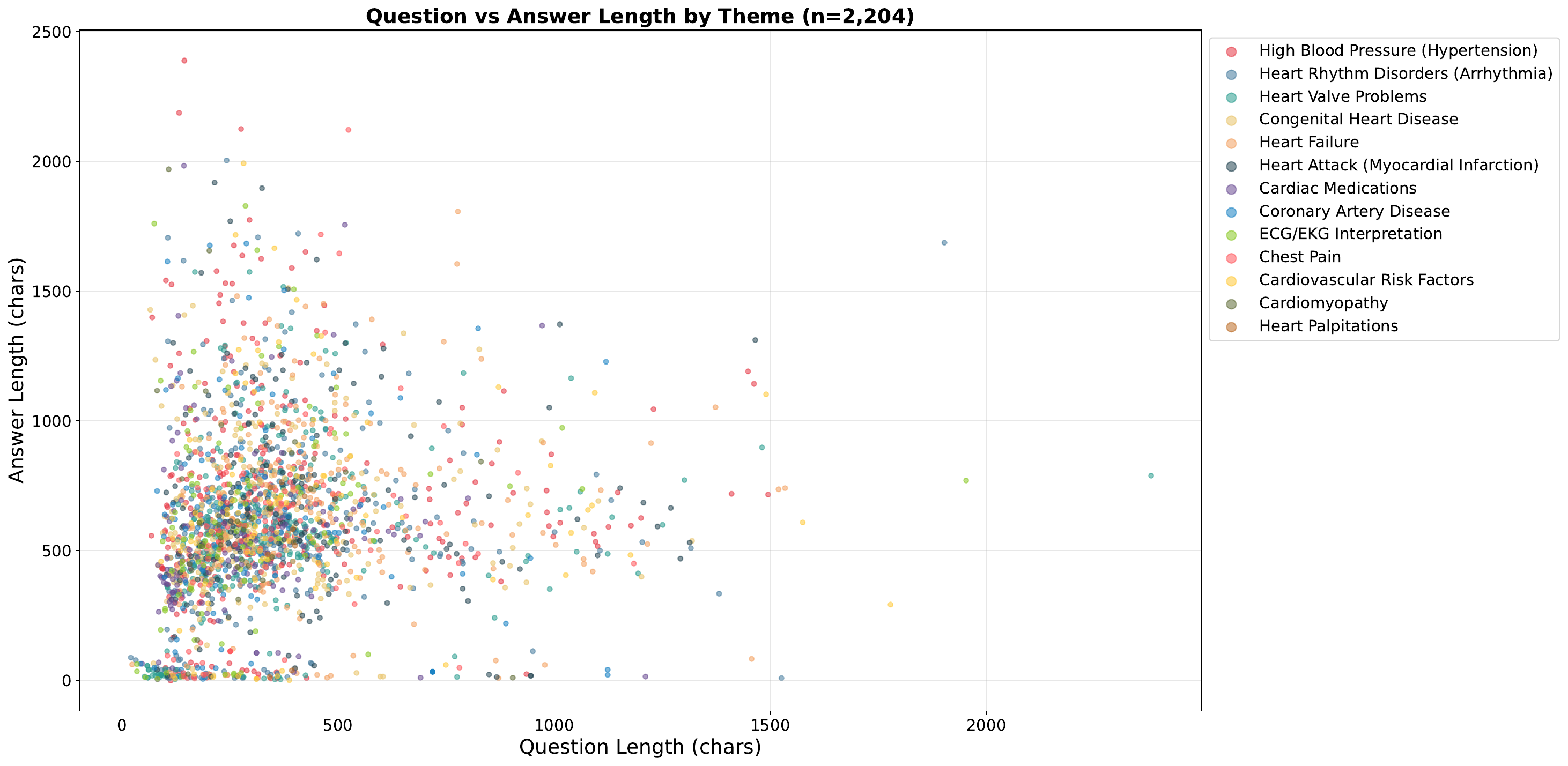}
  \suppFigCaption{Question length vs.\ answer length by theme}{Scatter plot colored by theme showing the positive correlation between question complexity and answer length.}
  \label{suppfig:question_answer_scatter}
\end{figure}

\FloatBarrier

\begin{figure}[h]
  \centering
  \includegraphics[width=\textwidth]{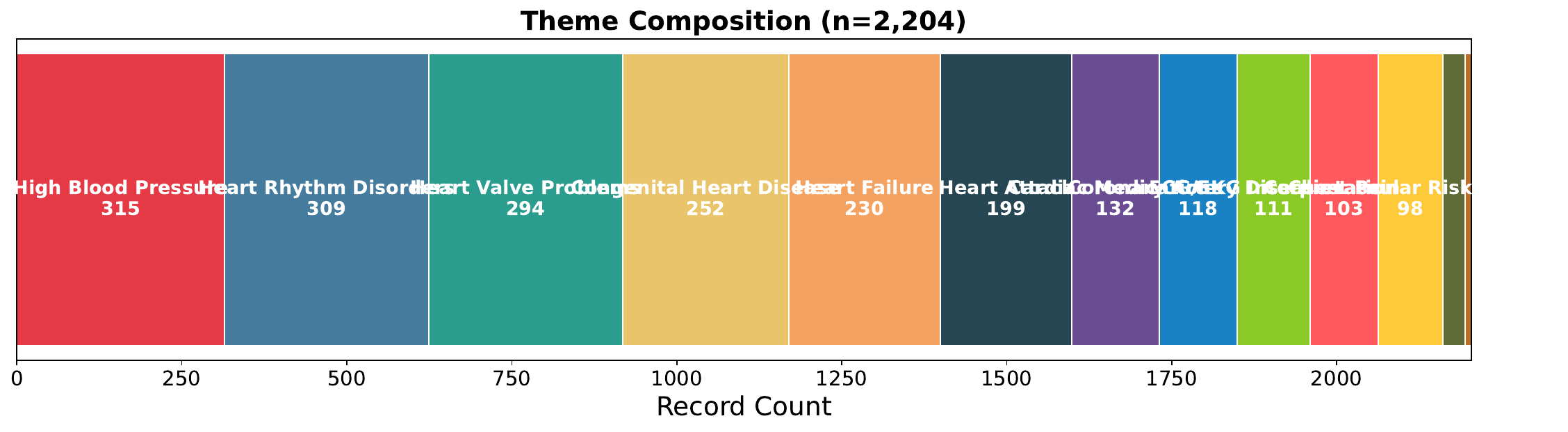}
  \suppFigCaption{Theme composition across the dataset}{Stacked bar chart showing relative composition of heart-related themes, highlighting the inter-theme imbalance.}
  \label{suppfig:theme_composition}
\end{figure}

\FloatBarrier

\section{Supplementary Note 11: Supplementary Evaluation Figures}
\label{supp:sec:eval_results}

These figures complement the main results (Table~2 and Fig.~4) with alternative metric visualizations.

\begin{figure}[h]
  \centering
  \includegraphics[width=\textwidth]{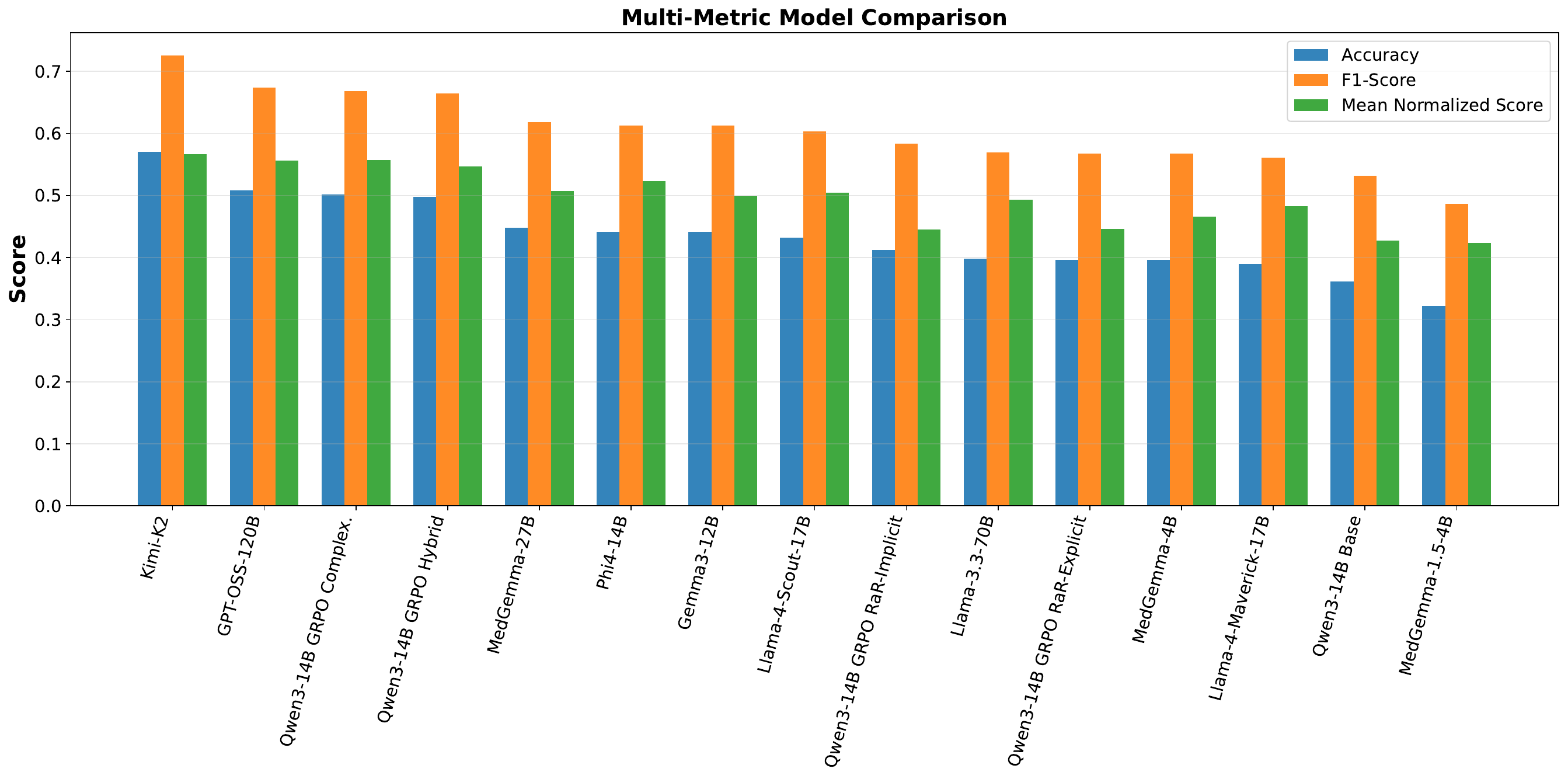}
  \suppFigCaption{Multi-metric comparison}{Grouped bar chart of accuracy, F1 score, and mean normalized score for the primary benchmark models.}
  \label{suppfig:multi_metric}
\end{figure}

\FloatBarrier

\begin{figure}[h]
  \centering
  \includegraphics[width=\textwidth]{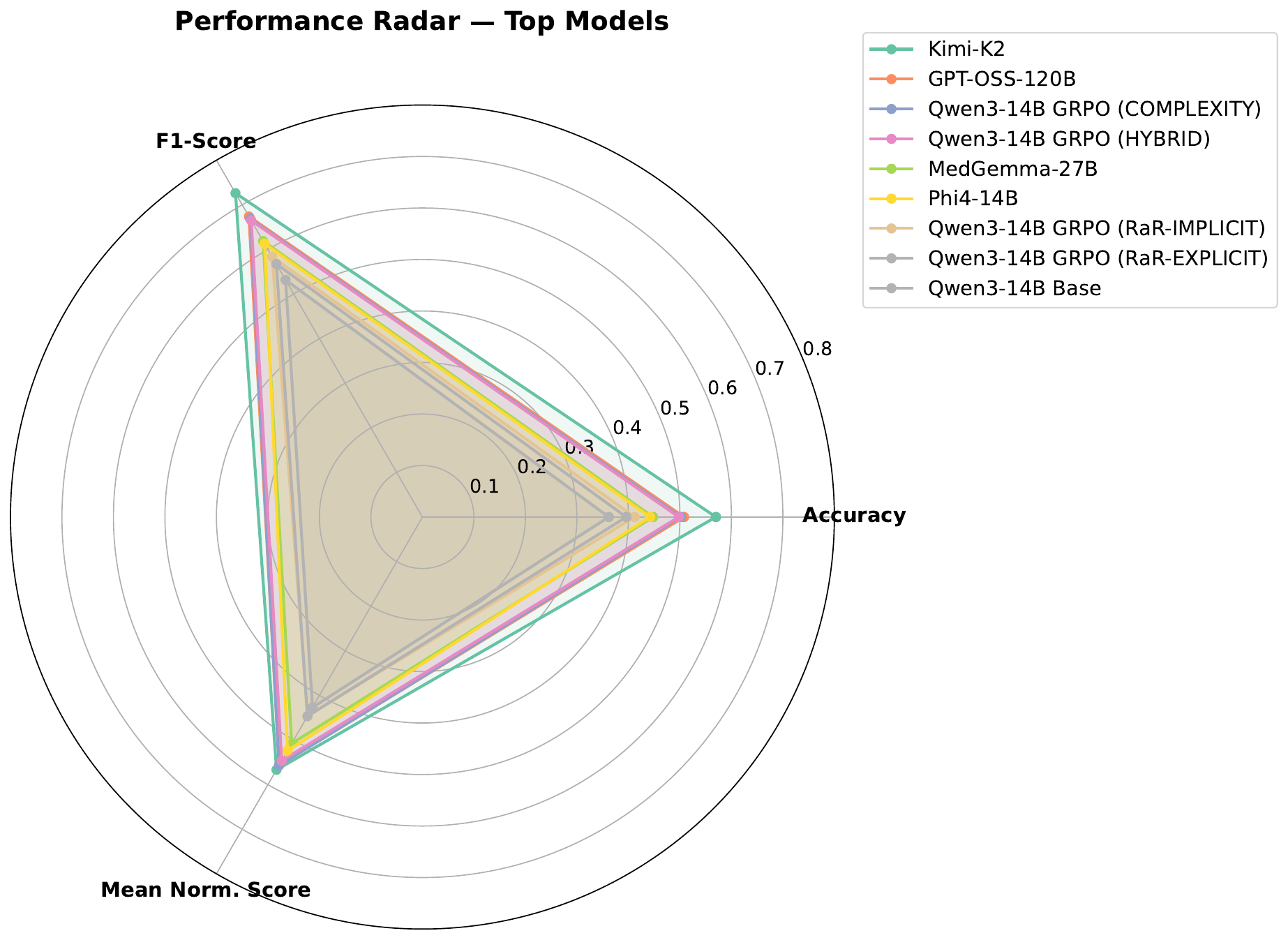}
  \suppFigCaption{Performance radar chart}{Radar chart showing accuracy, F1 score, and mean normalized score for the selected benchmark models. GRPO (COMPLEXITY) and GRPO (HYBRID) closely approach the coverage area of GPT-OSS-120B despite being ${\sim}9\times$ smaller.}
  \label{suppfig:radar}
\end{figure}

\FloatBarrier

\begin{figure}[h]
  \centering
  \includegraphics[width=\textwidth]{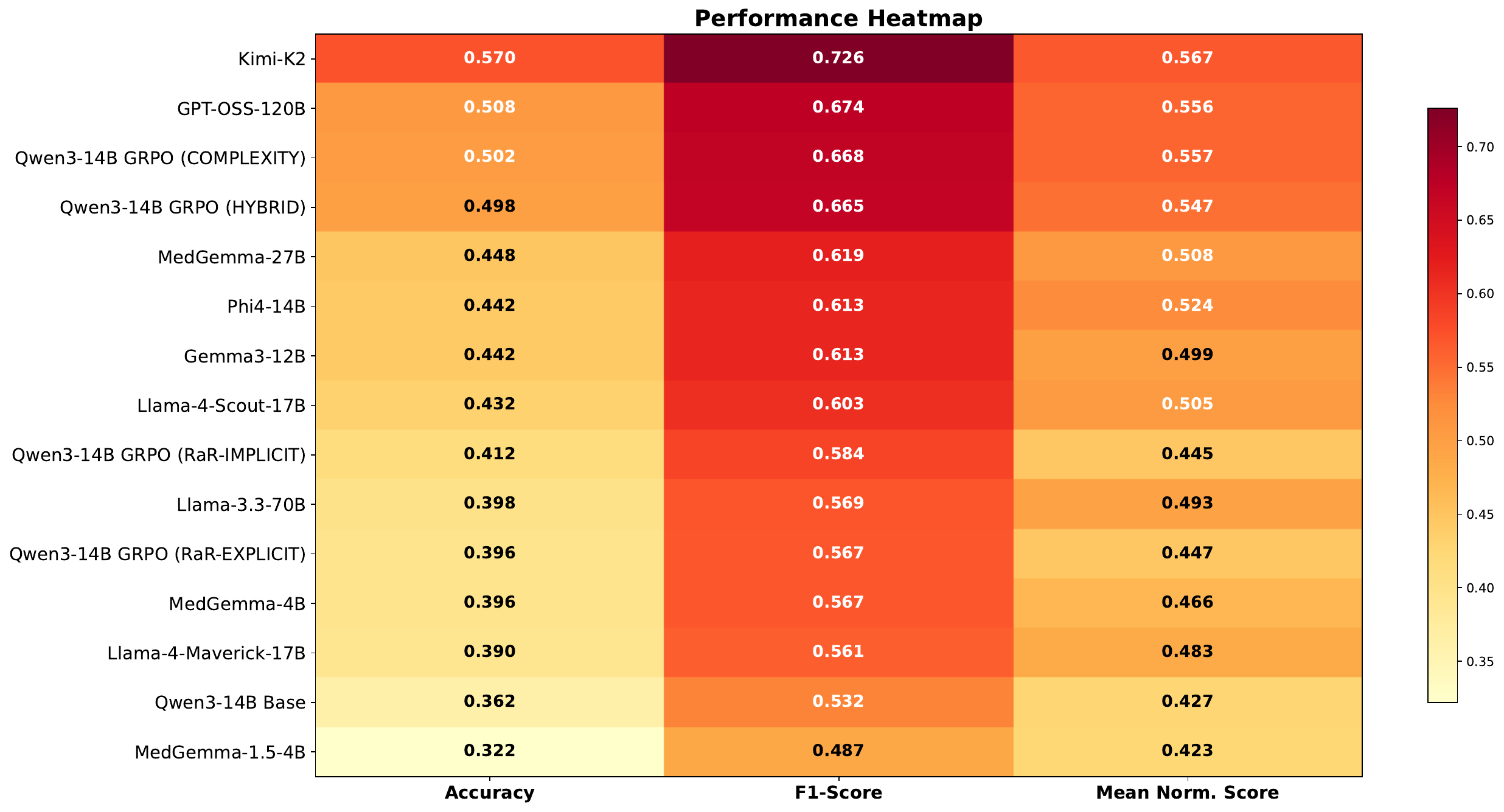}
  \suppFigCaption{Performance heatmap}{Heatmap of accuracy, F1 score, and mean normalized score for the primary benchmark models. The GRPO variants cluster with top-tier external models.}
  \label{suppfig:heatmap}
\end{figure}

\FloatBarrier

\begin{figure}[h]
  \centering
  \includegraphics[width=\textwidth]{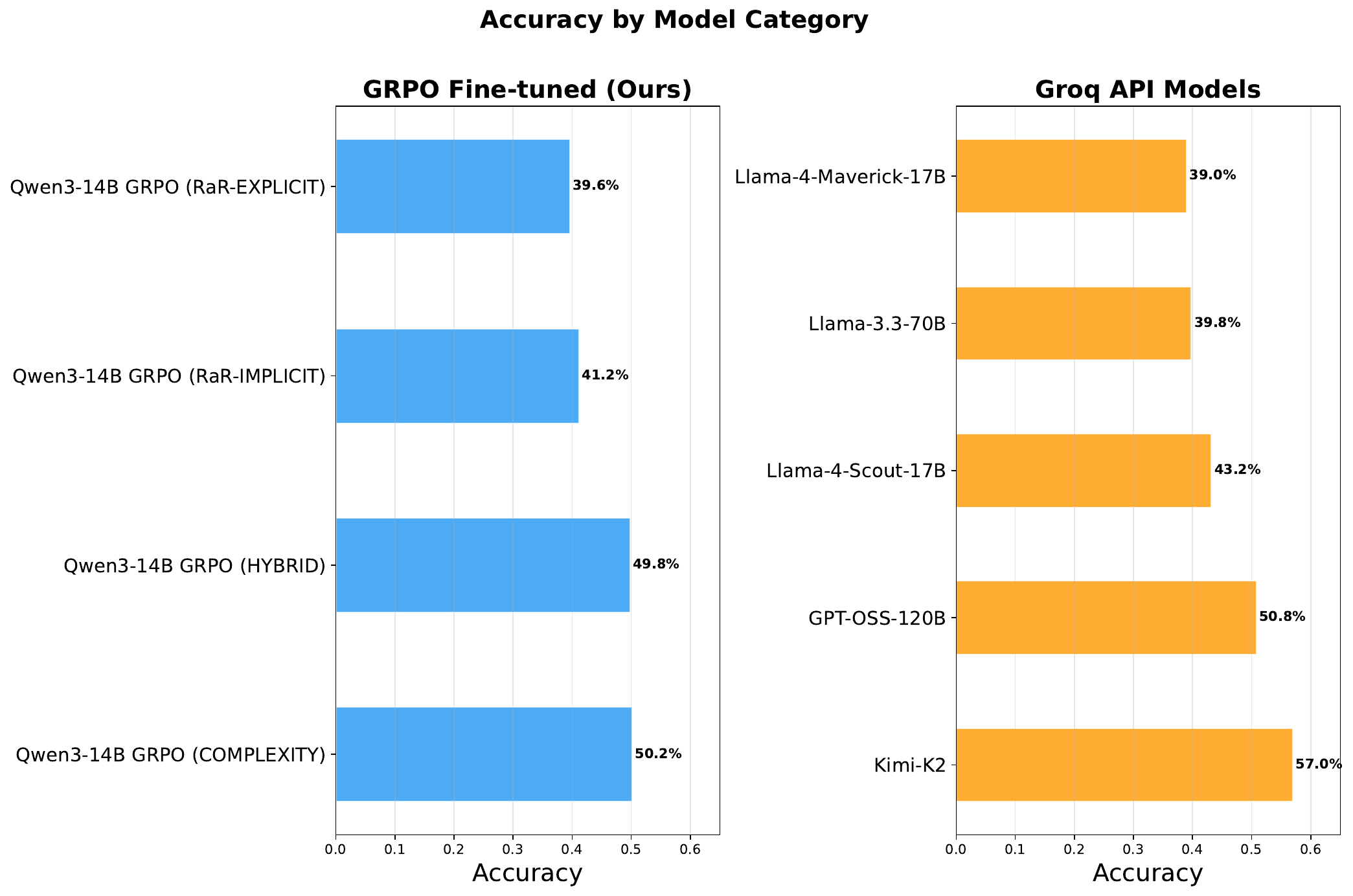}
  \suppFigCaption{Accuracy by model category}{Accuracy for the GRPO variants and Groq API models, shown in separate panels.}
  \label{suppfig:category_comparison}
\end{figure}

\FloatBarrier

\begin{figure}[h]
  \centering
  \includegraphics[width=\textwidth]{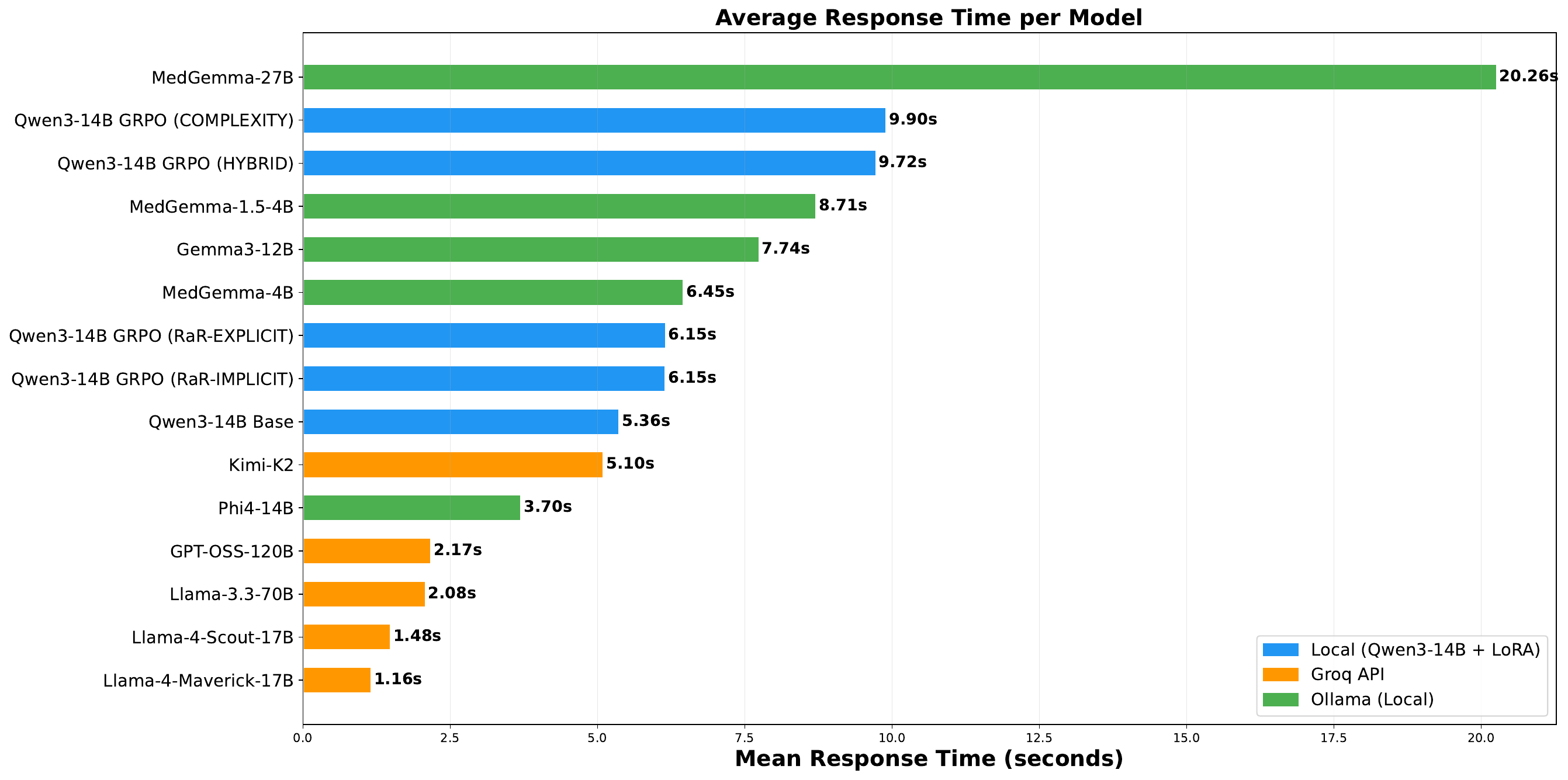}
  \suppFigCaption{Response time comparison}{Mean response times across the primary benchmark models. Locally deployed GRPO variants offer lower latency than MedGemma-27B while achieving higher accuracy.}
  \label{suppfig:response_times}
\end{figure}

\FloatBarrier

\section{Supplementary Note 12: Error Analysis}
\label{supp:sec:errors}

Manual inspection of evaluation errors on the HealthBench heart subset reveals five recurring failure categories, ordered by frequency:

\begin{enumerate}[leftmargin=*]
  \item \textbf{Incomplete criterion coverage} (most common): The model addresses the primary concern but omits secondary criteria---e.g., correctly recommending emergency evaluation for chest pain but failing to enumerate specific red-flag symptoms listed in the rubric.

  \item \textbf{Over-generalization}: The model provides generic cardiac advice rather than engaging with specific clinical details in the prompt (e.g., drug interactions, dosing considerations). This occurs more frequently on complex multi-part questions.

  \item \textbf{Negative criterion violations} (rare but high-impact): The model triggers a negative rubric criterion by providing contraindicated advice---e.g., recommending exercise without safety caveats for unstable conditions, or omitting relevant contraindications for suggested medications.

  \item \textbf{Reasoning--answer misalignment}: The reasoning trace identifies correct clinical considerations, but the final \texttt{<SOLUTION>} segment fails to incorporate all of them, suggesting the model's summarization ability lags behind its reasoning.

  \item \textbf{Format compliance failures}: Malformed tags or empty solution sections cause incomplete extraction. This primarily affects the base model before GRPO; after training, format compliance is near-universal.
\end{enumerate}

Error rates are not uniform across themes. Multidrug management themes (anticoagulation, heart failure with comorbidities) have higher error rates than simple clinical themes (hypertension, lifestyle advice). This expected pattern---more criteria per prompt means more opportunities to fail---is partially addressed by the Complexity-aware reward's logarithmic bonus for harder prompts.

After GRPO training, the error distribution shifts: the base model's dominant failure mode is rubric misalignment (factually correct but not criterion-targeted responses), while the GRPO models' errors concentrate on partial criterion coverage (criterion-aware but incomplete responses). This qualitative shift confirms that the reward signal teaches the model to attend to rubric-relevant aspects.

\subsection{Judge agreement}

To quantify evaluation noise, 50 responses were re-evaluated with a second independent judge call (same model and parameters). Criterion-level concordance exceeded 95\%, with disagreements concentrated on borderline criteria where the response partially addressed the criterion text. This supports the reliability of the single-call evaluation protocol.

\section{Supplementary Note 13: Checkpoint Strategy and Reproducibility}
\label{supp:sec:checkpoints}

\begin{itemize}[leftmargin=*]
\item \textbf{SFT checkpoints}: Saved to \texttt{sft\_saved\_lora/} after SFT completes. On subsequent runs, if this checkpoint exists, SFT is skipped and the adapter is loaded directly, enabling iterative experimentation with different GRPO reward functions without re-running the warm start.
\item \textbf{GRPO checkpoints}: Saved to \texttt{grpo\_saved\_lora\_\{reward\_type\}/} after GRPO completes, with intermediate checkpoints every 10 steps for post-hoc training dynamics analysis.
\item \textbf{Checkpoint verification}: After saving, the pipeline verifies that not all LoRA tensor weights are zero using \texttt{safetensors}. If all tensors are zero, a \texttt{RuntimeError} is raised to prevent using a failed checkpoint for evaluation.
\end{itemize}

\section{Supplementary Note 14: Software and Hardware}
\label{supp:sec:software}

\begin{table}[h]
\centering
\small
\begin{tabular}{l l}
\toprule
Component & Version / Specification \\
\midrule
GPU & NVIDIA RTX 6000 Pro (96\,GB VRAM) \\
Framework & PyTorch (with CUDA) \\
Training library & Unsloth + TRL (SFTTrainer, GRPOTrainer) \\
Inference engine & vLLM (integrated via Unsloth fast\_generate) \\
LoRA library & PEFT (Parameter-Efficient Fine-Tuning) \\
Quantization & bitsandbytes (4-bit NF4) \\
Judge API & Groq API (GPT-OSS-120B) \\
\bottomrule
\end{tabular}
\suppTableCaption{Software and hardware stack}{All training and evaluation was conducted on a single GPU.}
\label{tab:software}
\end{table}

All experiments were conducted on a single NVIDIA RTX 6000 Pro GPU with 96\,GB VRAM. 4-bit quantization, LoRA rank=16 and gradient checkpointing allow the full GRPO loop (including vLLM generation of 6 completions per prompt) to fit in the memory of a single GPU. This shows the feasibility of rubric-based RL for medical LLMs without needing multi-GPU clusters.

\section{Supplementary Note 15: AI Use Declaration}
\label{supp:sec:ai_use}

Generative AI assisted language editing and the flow of selected passages during manuscript preparation. The authors developed the research ideas, algorithms, experimental design, and scientific interpretations. The figures and reported numerical results were not generated or changed by these language editing tools. The authors retain responsibility for the accuracy and final wording of the manuscript.

Large language models were used as part of the training and evaluation workflow. MedGemma-27B generated synthetic reasoning traces for supervised fine-tuning examples, as described in Supplementary Note~3. GPT-OSS-120B served as the criterion-level judge during reward computation and evaluation, as described in Supplementary Note~1 and in the Methods section of the main paper. The authors designed the study, wrote and reviewed the manuscript, verified the analyses, and take full responsibility for the final content.